\documentclass[10pt,twocolumn,letterpaper]{article}

\usepackage{cvpr}              

\usepackage{graphicx}
\usepackage{amsmath}
\usepackage{amssymb}
\usepackage{booktabs}

\usepackage{algorithm}
\usepackage{algorithmic}
\usepackage{multirow}
\usepackage{tikz}
\usepackage[normalem]{ulem}
\useunder{\uline}{\ul}{}

\usepackage[pagebackref,breaklinks,colorlinks]{hyperref}

\usepackage[capitalize]{cleveref}

\usepackage{enumitem}

\begin{document}

\title{3C: Confidence-Guided Clustering and Contrastive Learning for Unsupervised Person Re-Identification}

\author{
Mingxiao Zheng $^{1}$\hspace{0.5cm}
Yanpeng Qu$^{1}$\hspace{0.5cm}
Changjing Shang$^{2}$\hspace{0.5cm}
Longzhi Yang$^{3}$\hspace{0.5cm}
Qiang Shen$^{2}$\\
${}^{1}$Dalian Maritime University\hspace{1cm}
${}^{2}$Aberystwyth University\hspace{1cm}
${}^{3}$Northumbria University
\vspace{-20mm}
}
\maketitle

\begin{abstract}
Unsupervised person re-identification (Re-ID) aims to learn a feature network with cross-camera retrieval capability in unlabelled datasets. Although the pseudo-label based methods have achieved great progress in Re-ID, their performance in the complex scenario still needs to sharpen up.
In order to reduce potential misguidance, including feature bias, noise pseudo-labels and invalid hard samples, accumulated during the learning process, in this paper, a confidence-guided clustering and contrastive learning (3C) framework is proposed for unsupervised person Re-ID.
This 3C framework presents three confidence degrees.
\romannumeral1) In the clustering stage, the confidence of the discrepancy between samples and clusters is proposed to implement a harmonic discrepancy clustering algorithm (HDC). 
\romannumeral2) In the forward-propagation training stage, the confidence of the camera diversity of a cluster is evaluated via a novel camera information entropy (CIE). Then, the clusters with high CIE values will play leading roles in training the model. 
\romannumeral3) In the back-propagation training stage, the confidence of the hard sample in each cluster is designed and further used in a confidence integrated harmonic discrepancy (CHD), to select the informative sample for updating the memory in contrastive learning. Extensive experiments on three popular Re-ID benchmarks demonstrate the superiority of the proposed framework.
Particularly, the 3C framework achieves state-of-the-art results: 86.7\%/94.7\%, 45.3\%/73.1\% and 47.1\%/90.6\% in terms of mAP/Rank-1 accuracy on Market-1501, the complex datasets MSMT17 and VeRi-776, respectively.
Code is available at https://github.com/stone5265/3C-reid.
\end{abstract}

\section{Introduction}
Person re-identification (Re-ID) aims to retrieve target pedestrian across non-overlapping camera views. It is a challenging task due to the variance of capturing angle, camera resolution and light condition. In practice, the expansive cost of annotating cross-camera identity labels promotes the role of unsupervised person Re-ID models in unknown and novel scenarios. The unsupervised person Re-ID task is mainly conducted from two perspectives: unsupervised domain adaptation (UDA) \cite{ECCV20JVTC,ICLR20MMT,NPIS20SpCL} and purely unsupervised learning (USL) \cite{AAAI19BUC,CVPR20SSL,CVPR20HCT}, depending on whether an external labelled source domain is used or not. 
UDA tasks are built in an annotation-known source domain via generative networks, such as GAN \cite{2014GAN},
to reduce the gap between it and the annotation-unknown target domain \cite{CVPR19PAUL,ECCV20DG-Net++,TPAMI20GPP}.
USL tasks are mainly implemented as pseudo-label based (PLB) models, which are trained under the supervision of pseudo-labels generated by either clustering algorithms \cite{TOMM18PUL,ICLR20MMT,NPIS20SpCL} or measuring similarities \cite{AAAI19BUC,CVPR20SSL,CVPR20HCT}. Moreover, generative networks also can be used to increase the amount of data for USL \cite{CVPR21GCL}.
Compared to UDA, USL utilises less information so that USL is more challenging but more practicable. In particular, using PLB models with clustering methods has become one of the mainstream learning paradigms for the unsupervised person Re-ID task. Therefore, this paper focuses on PLB models for better potential.

In the training process of PLB models, as a benchmark, SpCL \cite{NPIS20SpCL} employs contrastive learning to replace metric learning loss functions \cite{2017Tri,ICCV19PAST,CVPR20HCT,ICLR20MMT} which are inherited from supervised Re-ID tasks. 
To further mitigate the inadequacy of contrastive learning in the training stage framework, 
in GCL \cite{CVPR21GCL}, GAN was also used to augment data for contrastive learning in the USL task. 
The cluster-level memory, introduced in Cluster-Contrast \cite{ACCV22ClusterContrast}, also inspired number of subsequent attention to memory update approaches.
ICE \cite{ICCV21ICE} proposed to mine hardest positive samples in mini-batch for hard instance contrast, to reduce intra-cluster variance.
PPLR \cite{CVPR22PPLR} designed a part-based pseudo-label refinement framework, in which different local parts are utilised to smooth the global pseudo-labels.
CACL \cite{TIP22CACL} enhanced feature representation using the Siamese network in a designed asymmetric contrast learning framework.
ISE \cite{CVPR22ISE} generated support samples around the cluster boundaries to calibrate the incorrect learning direction caused by noise.

The clustering stage can be done by partition-based clustering algorithms, such as $K$\!-means \cite{1967KMeans}). 
However, the outcome of the $K$\!-means based methods \cite{TOMM18PUL,ICLR20MMT} is sensitive to the pseudo-label noise, which is from two sources. One is the weakly discriminative power of features learned in early iterations. Such features can hardly depict the differences between clusters (IDs), thus will mislead the model training. Another source happens in the complex scenario, where the distinction between images due to the camera bias may be more significant than that due to the actual IDs.

Due to above defects of $K$\!-means, density-based methods, such as DBSCAN \cite{1996DBSCAN} have drawn more attention in the clustering stage of PLB models \cite{ICCV19PAST,ICCV19SSG,NPIS20SpCL}.
Yet the work in \cite{TIP23RTMem} remarked that the use of contrastive learning in PLB models assumes a spherical data distribution, which is incompatible with the manifold assumption of DBSCAN. 
Moreover, in PLB models, the hyper-parameters of DBSCAN may not constantly fit the changing feature space during the training process. This discordance causes DBSCAN to tend to generate excessive sub-clusters (IDs) to fit one complex cluster (ID). 
Therefore, the unfashionable partition-based clustering algorithm is revisited for the Re-ID task in this paper.

To cope with the noise caused by weakly discriminative power of features and camera bias, this paper proposes a Confidence-guided Clustering and Contrastive learning (3C) framework. This 3C framework investigates three confidence degrees to improve the clustering, the forward-propagation training and the back-propagation training stages for PLB models, respectively.

In the clustering stage, the 3C framework develops the fuzzy-rough harmonic discrepancy clustering algorithm (HDC) \cite{TFS23HDC} with the confidence of the discrepancy between samples and clusters in partition-based clustering. 
Moreover, HDC also uses a peripheral object detection strategy to comfort the impact of the anomalies associated with the decentralised distribution of the cluster.

\begin{figure}[!htb]
    \centering
    \begin{subfigure}{0.19\linewidth}
        \includegraphics[width=\linewidth]{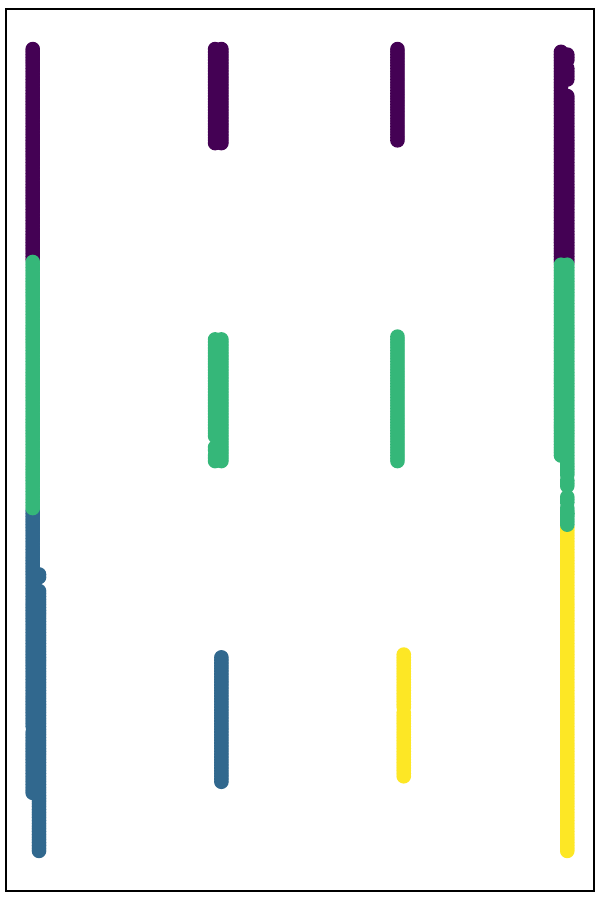}
        \caption{$K$\!-means}
        \label{fig:kmeans_example}
    \end{subfigure}
    \hfil
    \begin{subfigure}{0.19\linewidth}
        \includegraphics[width=\linewidth]{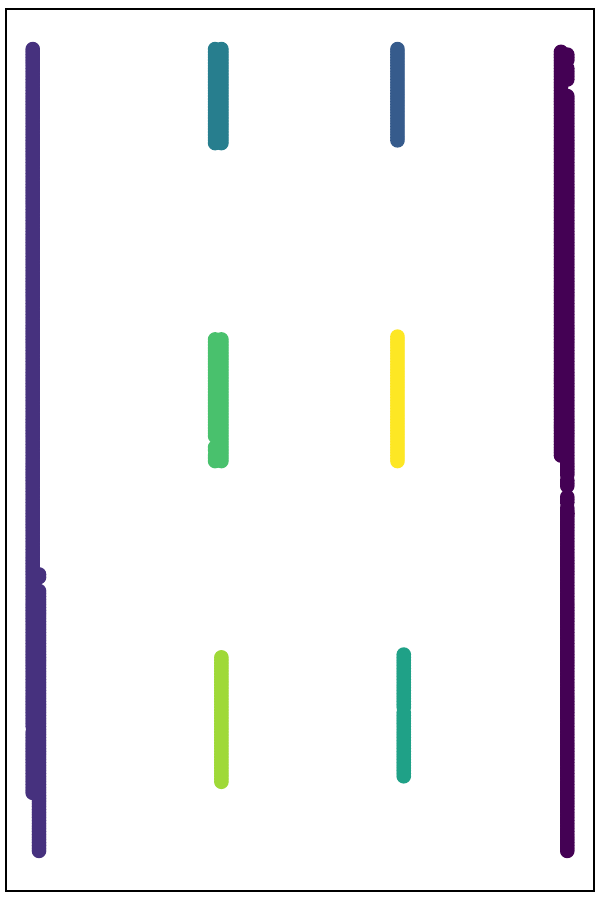}
        \caption{DBSCAN}
        \label{fig:dbscan_example}
    \end{subfigure}
    \hfil
    \begin{subfigure}{0.19\linewidth}
        \includegraphics[width=\linewidth]{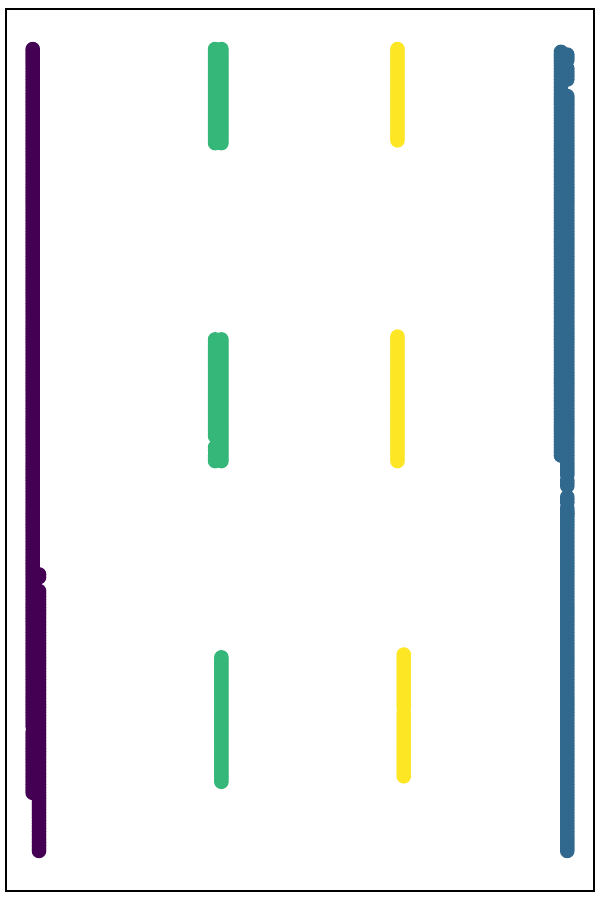}
        \caption{HDC}
        \label{fig:hdc_example}
    \end{subfigure}
    \caption{Comparison among $K$\!-means, DBSCAN and HDC.}
    \label{fig:clustering_examples}
\end{figure}

As shown in Fig. \ref{fig:clustering_examples}, experimental results of the road line segmentation tasks in \cite{TFS23HDC} are used as a sketch map to illustrate the comparison between $K$\!-means, DBSCAN and HDC.

Fig. \ref{fig:kmeans_example} shows that since the respective clusters represent the four road lines (two solid lines and two dashed lines) are close to each other and not spherical, the cluster centroids in $K$\!-means are tugged away by the neighbour clusters. These shifted centroids will lead to incorrect clusters, namely pseudo-label noise for Re-ID.  
Although DBSCAN successfully identifies the density regions, as shown in Fig. \ref{fig:dbscan_example}, it divides each dashed line into three independent sub-clusters, mistakenly. These secondary clusters indicate the challenges faced by DBSCAN, when clusters have different densities, as the fixed parameters might not suit all clusters. Such shortcoming will results in excessive pseudo-labels which are more than the real IDs for the Re-ID task.
Meanwhile, the correct clustering results of HDC shown in Fig. \ref{fig:hdc_example} demonstrate its superior clustering performance over others.

Since Re-ID is a cross-camera retrieval task, sources of data in each cluster are expected to be diverse. 
In fact, more cross-camera information can provide greater angle and scene variations for the ID images. The learning value of these images is higher, accordingly.
However, in the early learning stage, due to the low discriminative ability of features, the inter-camera similarity between images belong to the same ID may be lower than the inter-ID similarity between images belong to the same camera. This bias may result in a cluster consists of the images almost from a single camera.
To address this drawback, in the forward-propagation training stage, the proposed 3C framework presents a camera information entropy (CIE) to evaluate the confidence of the camera diversity for a cluster. The high/low CIE of a cluster indicates that the data in this cluster come from a large/little amount of sources.
For each cluster, the value of CIE will be converted into the weight for the corresponding infoNCE loss function in the contrastive learning process. In so doing, the learning rates of the clusters with low CIE will be transferred to those with high CIE.
As a result, the clusters with high CIE will play an important role in the contrastive learning loss to dominate the training stage.

In the back-propagation training stage, the infoNCE loss function in contrastive learning needs a memory to record and update the centroid of each cluster. The popular hard sampling strategy \cite{CVPR22ISE} applies the sample farthest from its cluster centroid to update the memory. However, such hard sampling method may fail for complex large-scale datasets \cite{ACCV22ClusterContrast,CVPR22ISE}, because the selected samples have the potential to be noise in the complex scenario.
To select a trustworthy sample for memory updating, the proposed 3C framework designs a confidence of the hard sample in a cluster,
from the perspective of the affiliation of a sample to its camera. Based on this confidence, a confidence integrated harmonic discrepancy (CHD) is further proposed to selected the valuable hard sample, which is  far from its cluster centroid and also close to its camera centre.

In summary, the main contributions of the proposed 3C framework can be concluded as follows.
\begin{itemize}
    \item In the clustering stage, the confidence of the discrepancy between samples and clusters is proposed to implement the HDC algorithm. The novel partition-based clustering algorithm can set a proper number of clusters for Re-ID, address the impact of noise on cluster centroids, and fit the assumption of spherical data distribution in the memory update strategy.
    \item In the forward-propagation training stage, the confidence of the camera diversity for a cluster is measured by CIE. By using CIE, a weighted infoNCE loss is proposed to reward/punish the clusters with high/low CIE, via large/small learning rates.
    \item In the back-propagation training stage, a confidence of the hard sample is proposed. The associated CHD metric can efficiently locate the reliable hard sample and the update direction of the memory, for the sake of both clusters and cameras.
    \item Extensive experiments on three popular Re-ID benchmark datasets demonstrate that the proposed 3C framework significantly outperforms existing state-of-the-art unsupervised person Re-ID methods.
\end{itemize}

The remainder of this paper is structured as follows: 
Section \ref{sec:related_work} reviews clustering, camera-aware and memory update strategies in unsupervised person Re-ID.
Section \ref{sec:proposed_approach} details the proposed approach.
Section \ref{sec:experiments} discusses the experiments on three popular Re-ID benchmark datasets.
The paper is concluded in Section \ref{sec:conclusion} with a discussion of further work.

\section{Related Work}
\label{sec:related_work}
\subsection{Clustering in Unsupervised Person Re-ID}
The early work PUL \cite{TOMM18PUL} relied on $K$\!-means to generate pseudo-labels and select reliable samples close to the cluster centroids for training. However, as a classical partition-based clustering algorithm, $K$\!-means is susceptible to the outliers and may impair the quality of generated pseudo-labels.
Therefore, BUC \cite{AAAI19BUC} merged clusters bottom-up using the sample similarity to incrementally generate more stable pseudo-labels. HCT \cite{CVPR20HCT} utilised hierarchical clustering to further enhance the reliability of pseudo-labels.
Furthermore, in  \cite{ICCV19PAST,ICCV19SSG}, density-based clustering methods, such as DBSCAN, came into use as the baseline method in the clustering paradigm of the PLB model.

LP \cite{TIP23LP} generated multiple clustering results via globally and locally pseudo-labels for complementary, and further reduced noise using the offline knowledge distillation.
Likewise, AdaMG \cite{TCSVT23AdaMG} generated multiple sets of pseudo-labels using different clustering settings, and adopted the online knowledge distillation.
DCCT \cite{TCSVT23DCCT} utilised a pair of identically structure networks for generating  pseudo-labels separately, which are used to co-teaching.

\subsection{Camera-aware Unsupervised Person Re-ID}
Since Re-ID is a cross-camera retrieval task, camera labels are employed as a supplement to data learning in some Re-ID works.

SSL \cite{CVPR20SSL} introduced a regularisation term to the instance similarity in the same camera, to encourage cross-camera instance similarity. Similar to it, the proposed 3C framework also exploits cross-camera behaviour.
IICS \cite{CVPR21IICS} decomposed the sample similarity from the intra-camera and inter-camera perspectives, to avoid feature bias from different cameras.
CAP \cite{AAAI21CAP} combined camera labels to further subdivide the clustering results, to generate camera-aware proxies instead of camera-agnostic clustering centroids.
CCL \cite{TCSVT23CCL} designed a time-based camera contrastive learning component to avoid a surge in the number of proxies caused by CAP.

\subsection{Memory update}
Since the feature network is gradually updated by learning process, the features of the proxy stored in the memory are shifted from the latest features. Therefore, the memory update strategy is crucial for the contrastive learning paradigm of the PLB model. 

RTMem \cite{TIP23RTMem} pointed out that current momentum-based memory update strategy is on the assumption that the spherical distribution of data, which is at odds with the assumption of used density clustering. Therefore, it presented a real-time memory update strategy, which is to randomly sample for override updating instead of momentum updating. By revisiting such inconsistency between training and clustering for the PLB model, this paper refocuses on the role of the partition-based clustering algorithm in the unsupervised person Re-ID task.

In order to prevent the memory features from being contaminated with noise pseudo-labels, AdaMG \cite{TCSVT23AdaMG} proposed a strategy to update the memory by measuring the confidence. It calculates the confidence of samples by balancing the distances of the sample-to-centroid between in the offline clustering result and in the online representation learning, and weights the samples for updating. Although this paper also designs a confidence term for memory update, the proposed approach will work on the hard sampling strategy.

With the help of the camera information, CCL \cite{TCSVT23CCL} constructed the camera centres per mini-batch for hard sampling update strategy to dilute the negative effect of noise. Different from CCL which uses camera centre directly for memory update, this paper constructs the camera centres in the confidence of samples, to select the reliable hard sample for updating.

DCMIP \cite{ICCV23DCMIP} and HCACE\cite{TMM24HCACE} used different memories to keep the direction of loss more stable. However, maintaining different memories will require extra computational complexity and space complexity.

\begin{figure*}
    \centering
    \includegraphics[width=16cm]{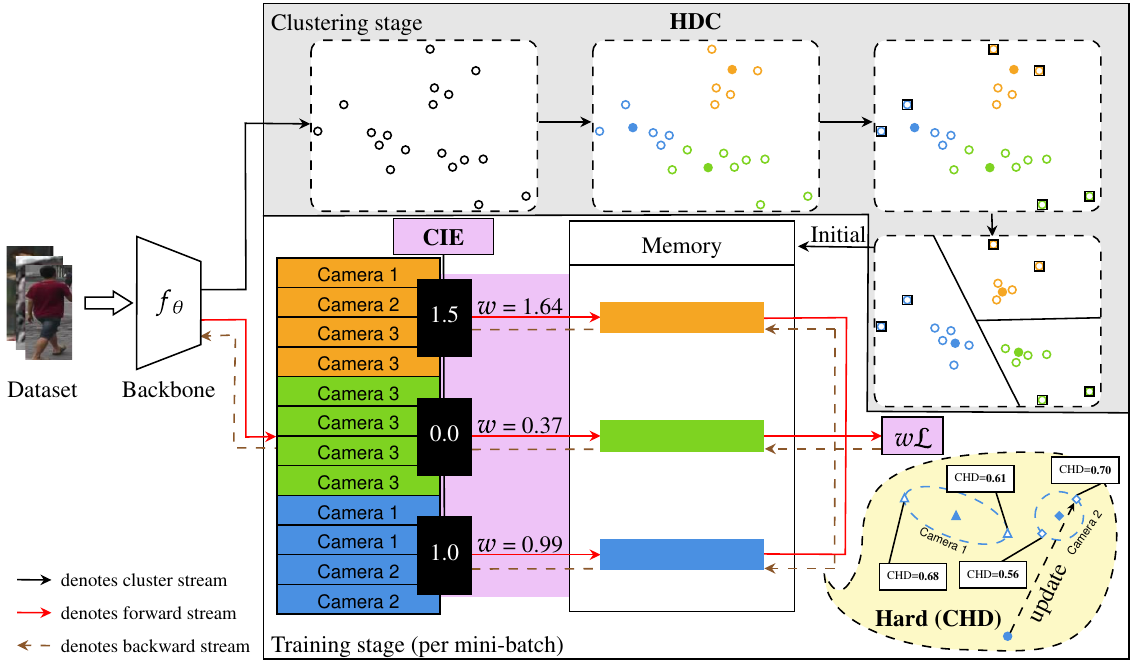}
    \caption{An overview on the proposed 3C framework.}
    \label{fig:framework}
\end{figure*}

\section{Proposed Approach}
\label{sec:proposed_approach}
This paper presents a confidence-guided clustering and contrastive learning (3C) framework for unsupervised person Re-ID.
The flowchart of this 3C framework is illustrated in Fig. \ref{fig:framework},  where the components highlighted in grey, pink and beige indicate three confidences designed to reduce the impact of the misdirection involved in the clustering, forward-propagation training and back-propagation training stages.

\subsection{Preliminary and Revisiting}
Given an unlabelled person Re-ID dataset $\mathcal{X} = \{x_i\}^{N}_{i=1}$, where $x_i$ is the $i$-th image and $N$ is the number of images. The goal of unsupervised person Re-ID is to learn a network $f_{\theta}$ to project a sample $x_i$ to a embedding feature $f_{\theta}(x_i) \in \mathbb{R}^D$, where $\theta$ is the parameters of the network and $D$ is the dimension of features. 

In the learning phase of the PLB model, most of recent methods \cite{NPIS20SpCL,ICCV21ICE,CVPR22PPLR,ACCV22ClusterContrast,TIP22CACL,CVPR22ISE,TIP23RTMem,TIP23LP,TCSVT23AdaMG,TCSVT23CCL,TCSVT23DCCT} perform a two-step iterative strategy:

\romannumeral1) Generate pseudo-labels $\mathcal{Y} = \{y_i\}^{N}_{i=1}$ of all training images via offline clustering algorithms. Here, $y_i \in \{-1,1,2,...,K\}$ and $K$ is the number of clusters.

\romannumeral2) Optimise the network with the grouped dataset $\mathbb{U} = \{x_i,y_{i}\}^{N^{\prime}}_{i=1}$, where $N^{\prime}$ is the number of images after discarding outliers of $y_i=-1$.

In the inference phase, the similarity between the query and each image in the gallery set is calculated to retrieve the images which belong to the same ID as query from the gallery.

Some recent studies on the PLB model \cite{ACCV22ClusterContrast,TIP22CACL,CVPR22ISE,TIP23RTMem,TIP23LP,TCSVT23AdaMG,TCSVT23CCL,TCSVT23DCCT} employ contrastive learning and the non-parametric classifier \cite{2018InstDisc} based on the infoNCE loss function. The unified formulation of infoNCE is defined on $\mathbb{U}$ as:
\begin{equation}
\label{eq:infoNCE}
\mathcal{L} = -\log\frac
{\exp(\boldsymbol{f}_i \cdot \boldsymbol{m}_+ / \tau)}
{\sum^{M}_{k=1} \exp(\boldsymbol{f}_i \cdot \boldsymbol{m}_k / \tau)},
\end{equation}
where $\tau$ is a temperature factor, $\boldsymbol{f}_i$ is the simple representation of $f_{\theta}(x_i)$ and $\boldsymbol{m}_k$ is the $k$-th proxy picked from the memory $\mathcal{M}$. Among the $M$ proxies, $\boldsymbol{m}_+$ is the proxy which $x_i$ belongs to. There are several types of $\mathcal{M}$:

\romannumeral1) If $\mathcal{M}$ stores the features of all samples \cite{TIP23RTMem}, $M$ equals to $N^{\prime}$ and $\boldsymbol{m}_+$ is each sample belonging to $y_i$-th cluster. 

\romannumeral2) If $\mathcal{M}$ stores the centroids of all clusters \cite{ACCV22ClusterContrast,TCSVT23CCL}, $M$ equals to $K$ and $\boldsymbol{m}_+$ is the centroid of $y_i$-th cluster.

\romannumeral3) Some methods \cite{AAAI21CAP,ICCV21ICE} utilise the camera labels to store centroids of all clusters split by different cameras.

Since both $\boldsymbol{f}_i$ and $\boldsymbol{m}_k$ are $\ell_2$-normalised, the cosine similarity $\boldsymbol{f}_i \cdot \boldsymbol{m}_k$ is adopted as the similarity score between features.

It is necessary to update the features stored in memory $\mathcal{M}$ during the back-propagation. The commonly used momentum update strategy \cite{2020MoCo} is as follows.
\begin{equation}
\label{eq:momentum_update}
    \boldsymbol{m}_+ \longleftarrow \alpha\boldsymbol{m}_+ + (1-\alpha)\boldsymbol{f}_i,
\end{equation}
where $\alpha \in [0,1]$ is the momentum  hyper-parameter that controls the update ratio for the memory. $\alpha = 0$ means dropping the original features in the memory. $\alpha = 1$ means giving up updating the memory. 

According to the sample $\boldsymbol{f}_i$ used for memory update, various strategies can be implemented via Eq. \eqref{eq:momentum_update}.

In \cite{ACCV22ClusterContrast}, all the samples in the mini-batch are used to update memory sequentially in the order of being sampled. This memory update strategy is notated as \textbf{Vanilla} in this paper.
Another concept selects the hard sample in the current mini-batch for each ID, according to certain rules for updating the corresponding centroid.
The popular \textbf{Hard} sampling strategy treats the sample furthest from the corresponding centroid as the most valuable one \cite{CVPR22ISE}.
In \cite{TCSVT23CCL}, the \textbf{Hard (TCCL)} strategy utilises the camera information in an ID (cluster) to generate the local camera centres with regard to this ID, and then selects the furthest camera centre as the hard sample to update the memory of this ID (cluster).

\subsection{Harmonic Discrepancy Clustering}
In our prior work \cite{TFS23HDC}, a fuzzy-rough harmonic discrepancy clustering algorithm (HDC) is proposed, which is thus featured by a powerful processing ability on complex data distribution leading to enhanced clustering performance. In this paper, HDC is further developed into a generalised version for large-scale datasets of the unsupervised person Re-ID task.

\subsubsection{Harmonic Discrepancy}
The similarity between sample $x_i$ and sample $x_j$ is defined as:
\begin{equation}
\label{eq:similarity}
sim(x_i,x_j) = \exp(-dist(x_i,x_j)),
\end{equation}
where $dist(x_i,x_j)$ denotes the distance between $x_i$ and $x_j$. The associated discrepancy  between $x_i$ and $x_j$ can be defined as:
\begin{equation}
\label{eq:discrepancy}
D(x_i,x_j) = 1 - sim(x_i,x_j).
\end{equation}

Moreover, the discrepancy of a sample $x_i$ to a cluster $C_k$ is defined as:
\begin{equation}
\label{eq:discrepancy2cluster}
D_k(x_i) = \max\limits_{x \in C_k}\{D(x_i,x)\},
\end{equation}
which can be deemed the max-link distance between $x_i$ and $C_k$.
However, the discrepancy of a sample to a cluster may undergo a high
probability of inaccuracy, if this cluster suffers from a decentralised distribution. In this case, the discrepancy can be improved by taking into account the confidence of the $x \in C_k$ in Eq. \eqref{eq:discrepancy2cluster}.

The centroid $c_k$ of cluster $C_k$ is defined as:
\begin{equation}
\label{eq:centroids}
c_k = \frac{\sum^{N}_{i=1}\boldsymbol{1}(y_i = k)f(x_i)}{\sum^{N}_{i=1}\boldsymbol{1}(y_i = k)},
\end{equation}
where $\boldsymbol{1}(\cdot)$ equals to 1/0 when the  argument is \textbf{True}/\textbf{False}.
Since the centroid represents the expectation of the samples belonging to this cluster, the confidence of the discrepancy in Eq. \eqref{eq:discrepancy2cluster} can be evaluated via the similarity of $x$ to $c_k$ as follows.
\begin{equation}
\label{eq:conf_x}
con\!f_k(x) = sim(x,c_k),
\end{equation}
in which a large value of $con\!f_k(x)$ indicates that $x\in C_k$ in  Eq. \eqref{eq:discrepancy2cluster} is trustworthy, thus leading to a reliable discrepancy.

By calculating the discrepancy of a sample $x_i$ to a cluster $C_k$ in Eq. \eqref{eq:discrepancy2cluster} along with maximising the confidence of $x \in C_k$ in Eq. \eqref{eq:conf_x}, a representative object $\tilde{x}_{ik}$ is sought to metric the harmonic discrepancy (HD) of $x_i$ to cluster $C_k$, which is designed as:
\begin{equation}
\label{eq:representative_sample}
\tilde{x}_{ik} = \mathop{\arg\max}\limits_{x \in C_k} \{
    \frac{2 \times D(x_i,x) \times sim(x,c_k)}{D(x_i,x) + sim(x,c_k)}
\}.
\end{equation}

The harmonic average of $D(x_i,x)$ and $sim(x,c_k)$ indicates that both of them are guaranteed to large values. 
With the support of Eq. \eqref{eq:representative_sample}, the HD value of $x_i$ to cluster $C_k$ can be expressed as:
\begin{equation}
\label{eq:harmonic_discrepancy}
H\!D_k(x_i) = D(x_i,\tilde{x}_{ik}).
\end{equation}

According to the definition of HD in Eq. \eqref{eq:harmonic_discrepancy}, the similarity of $x_i$ to $C_k$ can be obtained from the discrepancy expressing the degree of separating $x_i$ from other clusters, as follows.
\begin{equation}
\label{eq:hd3}
\mu_k(x_i) = \frac{1}{K-1}\sum\limits_{l \neq k}H\!D_l(x_i).
\end{equation}

Therefore, the label $y_i$ of sample $x_i$ is assigned as:
\begin{equation}
\label{eq:hd4}
y_i = \mathop{\arg\max}\limits_{1 \leq k \leq K}\{\mu_k(x_i)\}.
\end{equation}

Since the HDC clustering algorithm takes into account the discrepancy of a sample to cluster and the compactness of this cluster, which is deemed the confidence of the discrepancy, HDC can balance the advantages of partition-based clustering and density-based clustering.

\subsubsection{Peripheral Object Detection}
Despite the comprehensive strategy of HD to distinguish the degrees of data instances belonging to a cluster, its efficacy may still be compromised by the anomalies associated with the decentralised distribution of the cluster. Therefore, it is necessary to distinguish the core objects from the peripheral objects, so that the centroid updating will depend only the core.

To identify the peripheral objects of cluster $C_k$, the HD acceptance threshold of $x$ to the cluster $C_k$ is defined as:
\begin{equation}
\label{eq:threshold}
\epsilon_k = ave(H\!D_k(x)) + std(H\!D_k(x)),
\end{equation}
where $ave(H\!D_k(x))$ and $std(H\!D_k(x))$ represent the average and the standard derivation of $H\!D_k(x)$ for $\forall x \in C_k$, respectively. Given the threshold $\epsilon_k$, for any $x_i \in C_k$, if $H\!D_k(x_i)>\epsilon_k$, its affiliation with $C_k$ can be considered unreliable, naturally. Therefore, all such instances are deemed the peripheral objects of $C_k$; otherwise, they are labelled as a member of the core set. By omitting the peripheral
objects, if there is any, in line with Eq. \eqref{eq:centroids}, the likelihood of the occurrence of the offset
centroid is mitigated.

The process of HDC is summarised in Algorithm \ref{alg:hdc}. Specifically, in the lastest iteration, i.e., $iter=Max$, Steps 14-19 discard the peripheral objects of $y_i=-1$ as outliers, which won't be used in the following model training process.

\begin{algorithm}[ht]
\caption{\textbf{H}armonic \textbf{D}iscrepancy \textbf{C}lustering (HDC)}
\label{alg:hdc}
\begin{algorithmic}[1]
\REQUIRE ~~\\
$\mathcal{X}$, input space containing $N$ objects,\\
$K$, number of clusters,\\
$Max$, maximum iterations.
\ENSURE ~~\\
$\mathcal{Y}$, label vector of $\mathcal{X}$.
\STATE $C = \textbf{INIT\_CLUSTER}(\mathcal{X}, K)$;\
\FOR{$iter$ in $[1,Max]$}
    \FOR{$k$ in $[1,K]$}
        \STATE $c_k \leftarrow$  Eq. \eqref{eq:centroids};\
        \FOR{$i$ in $[1,N]$}
            \STATE $\tilde{x}_{ik} \leftarrow$ Eq. \eqref{eq:representative_sample};\
            \STATE $H\!D_k(x_i) \leftarrow$ Eq. \eqref{eq:harmonic_discrepancy};\
        \ENDFOR
    \ENDFOR
    \STATE $\mathcal{Y} \leftarrow$ Eqs. \eqref{eq:hd3} and \eqref{eq:hd4};\
    \FOR{$k$ in $[1,K]$}
        \STATE $C_k = \{x_i | y_i = k,i = 1,2,...,N\}$;\
        \STATE $\epsilon_k \leftarrow$ Eq. \eqref{eq:threshold};\
        \FORALL{$x_i \in C_k$}
            \IF{$H\!D_k(x_i) > \epsilon_k$}
                \STATE $y_i = -1$;\
                \STATE $C_k = C_k - \{x_i\}$;\
            \ENDIF
        \ENDFOR
    \ENDFOR
\ENDFOR
\RETURN $\mathcal{Y}$;
\end{algorithmic}
\end{algorithm}

\begin{figure*}[!htb]
    \centering
    \includegraphics[width=16cm]
    {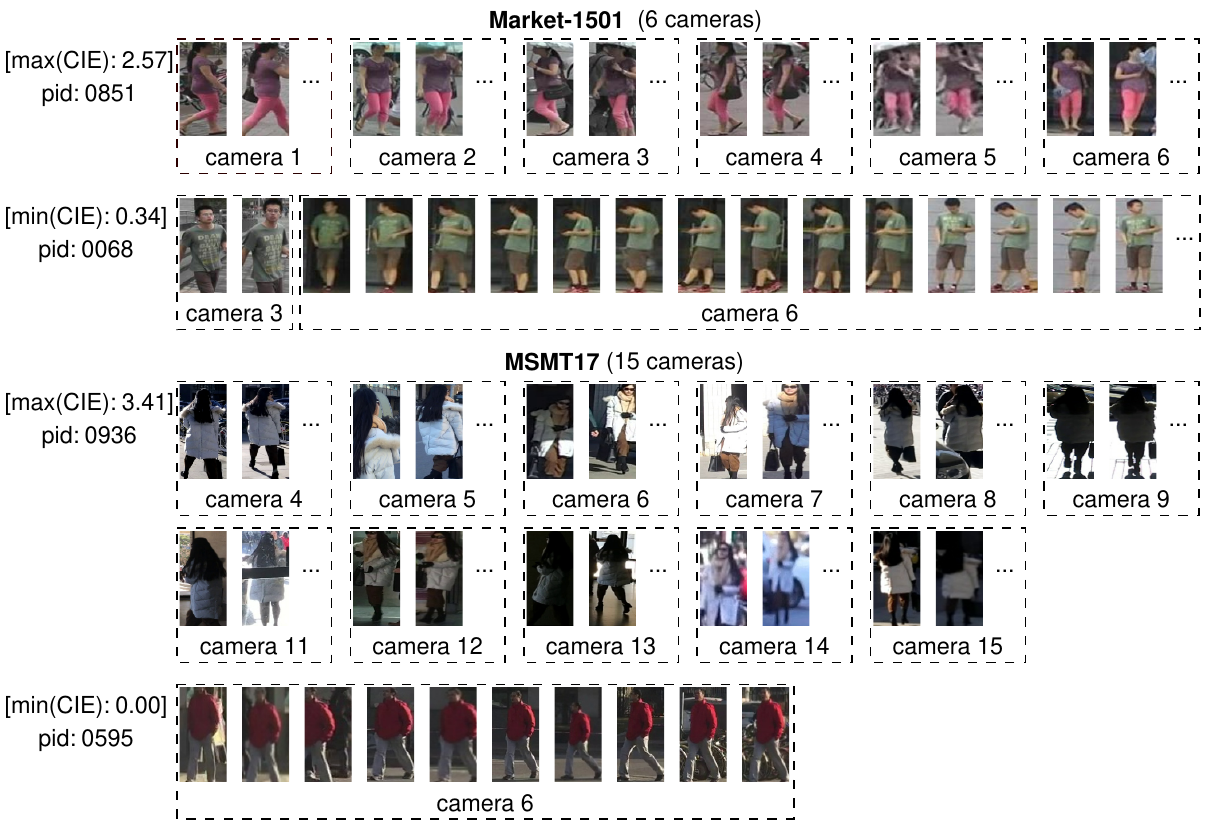}
    \caption{Illustration of CIE for real IDs.}
    \label{fig:CIE_illustration}
\end{figure*}

\subsection{Camera Information Entropy}
After the clustering stage, each resulting cluster will be coded with a pseudo-label.
Because Re-ID is a cross-camera task, each cluster is preferred to contain images from varied cameras. From this perspective, this paper further designs the confidence of the camera diversity for a cluster $C_k$, entitled camera information entropy (CIE), as follows. 
\begin{equation}
\label{eq:cam_info_entropy}
C\!I\!E(C_k) = \hspace{-1.2em}\sum_{Cam_i \in Cam}\hspace{-1.2em}-P(Cam_i|C_k)\log P(Cam_i|C_k),
\end{equation}
where
\begin{equation}
\label{eq:cam_info_entropy2}
P(Cam_i|C_k) = \frac{|Cam_i \cap C_k|}{|C_k|},
\end{equation}
and $Cam$ contains all camera sets, where $Cam_i$ is the set of images captured by the $i$-th camera.
For a cluster, a high value of CIE indicates a rich source of cameras, which is aligned with the cross-camera scenario of the Re-ID task. Thus, CIE can be used to gauge the confidence of the camera diversity for a cluster.

By using the $PK$ sampling method \cite{2017Tri}, the CIE of $P$ clusters $\{C'_k\}^P_{k=1}$ can be obtained per mini-batch, where $C'_k$ is the $k$-th cluster in this mini-batch. The resulting CIE of each cluster is further transformed into the weight of the infoNCE loss of this cluster as follows.
\begin{equation}
\label{eq:weighted_loss}
\mathcal{L}' = -w_+\log\frac
{\exp(\boldsymbol{f}_i \cdot \boldsymbol{m}_+ / \tau)}
{\sum^K_{k=1} \exp(\boldsymbol{f}_i \cdot \boldsymbol{m}_k / \tau)},
\end{equation}
where
\begin{equation}
\label{eq:camera_weight}
w_+ = \frac{\exp(C\!I\!E(C_+))}{\sum^P_{k=1} \exp(C\!I\!E(C'_k))} \times P.
\end{equation}

By using Eq. \eqref{eq:weighted_loss}, the contrastive learning process of the 3C framework favours the infoNCE loss associated to the cluster with  high CIE, over that with low CIE. This inductive bias conforms with the fact that the clusters with large CIE are full of camera sources, thus merit a high learning value for Re-ID. 
In the light of Eq. \eqref{eq:camera_weight}, the learning rate of the clusters (pseudo-labels) with low CIE will be transferred to the  clusters (pseudo-labels) with high CIE, which will dominate the training process of the 3C framework. Therefore, CIE may be deemed an indicator to predict the performance of an unsupervised Re-ID model.

The operations backgrounded in pink in Fig. \ref{fig:framework} schematically illustrate the role of CIE and the associates loss weight. Given three clusters sampled in a mini-batch, the green cluster reflects a false capture to the camera diversity because all images are from camera 3. The orange cluster, where the images are from cameras 1, 2 and 3, has been represented by the cross-camera information. 
Therefore, in line with Eq. \eqref{eq:weighted_loss} and Eq. \eqref{eq:camera_weight}, the loss weight with respect to the green/orange cluster is low/high, to reduce/enhance the impact of cluster on the model training process.

The role of CIE also can be observed in the benchmark datasets.
Fig. \ref{fig:CIE_illustration} shows the respective real IDs with the largest and smallest values of CIE in Market-1501 and MSMT17.
It can be seen that the images of ID 0851 contain all six cameras, thus enjoy the highest CIE value on Market-1501. And the images of ID 0068 are from only two cameras, mostly camera 6, thus gets the lowest CIE value. 
On the complex MSMT17, the images of ID 0595 all belong to the identical camera, which results in a 0 CIE. In particular, these images remain essentially unchanged in the same pose compared to the drastic change in the images of ID 0936 which has the largest CIE value. Thus, the role of ID 0936 is more significant  than that of ID 0595 for MSMT17.

\subsection{Confidence integrated Harmonic Discrepancy}

The hard sampling strategy chooses the sample furthest from the centroid to update the memory. However, such sample has the potential to be a noise, in the presence of a large number of noise pseudo-labels, which will mislead the direction of memory via Eq. \eqref{eq:momentum_update}. In this paper, a confidence of a sample to its camera is designed to evaluate the hard sample as follows.
\begin{equation}
\label{eq:conf_f}
   conf_{+}(\boldsymbol{f}_i)= sim(\boldsymbol{f}_i,\boldsymbol{O}_+),
\end{equation}
where $sim(\cdot)$ is the cosine similarity via normalisation. $\boldsymbol{O}_+$ is the camera centre corresponding to $\boldsymbol{f}_i$ in cluster $C_+$ and this mini-batch as following:
\begin{equation}
\label{eq:camera_centre}
    \boldsymbol{O}_+ = \frac{1}{|Cam_+ \cap C_+|}\sum_{\boldsymbol{f}_j \in Cam_+ \cap C_+}\hspace{-1.5em}\boldsymbol{f}_j.
\end{equation}

Under the guidance of a confidence integrated harmonic discrepancy (CHD), which is defined as:
\begin{equation}
\label{eq:hd_camera}
    C\!H\!D(\boldsymbol{m}_+,\boldsymbol{f}_i) = \frac{2 \times D(\boldsymbol{m}_+,\boldsymbol{f}_i) \times sim(\boldsymbol{f}_i,\boldsymbol{O}_+)}{D(\boldsymbol{m}_+,\boldsymbol{f}_i) + sim(\boldsymbol{f}_i,\boldsymbol{O}_+)},
\end{equation}
where $D(\cdot)$ is implemented by the cosine distance via normalisation.
The hard sample in this paper can be located by maximising CHD as follows.
\begin{equation}
\label{eq:memory_update_hd_hard}
    \boldsymbol{f}_{hard(C\!H\!D)} \longleftarrow \mathop{\arg\max}\limits_{\boldsymbol{f}_i \in C_+} \{ C\!H\!D(\boldsymbol{m}_+,\boldsymbol{f}_i) \}.
\end{equation}

According to Eq. \eqref{eq:momentum_update}, the proposed \textbf{Hard(CHD)} sampling method uses the sample selected via CHD to update the memory as follows.
\begin{equation}
\label{eq:memory_update_hard}
    \boldsymbol{m}_+ \longleftarrow \alpha\boldsymbol{m}_+ + (1-\alpha)\boldsymbol{f}_{hard(C\!H\!D)},
\end{equation}

By using the confidence \eqref{eq:conf_f}, a sample far from its cluster centroid is trustworthy, if it is also close to its camera centre.
Thus, by the CHD metric, the sample achieves the best performance on the tradeoff between hardship and confidence will be deemed the valuable hard sample and used to update memory. 

The sketch map backgrounded in beige in Fig. \ref{fig:framework} depicts the process of updating the centroid of the blue cluster stored in the memory. It can be seen that compared to the right camera, the left camera has samples more distant from the cluster centroid. Given the confidence in Eq. \eqref{eq:conf_f}, the largest CHD, used to update the memory, may be achieved by the sample belong to the right camera. 

\begin{figure}[!htb]
    \centering
    \begin{subfigure}{0.47\linewidth}
        \includegraphics[width=\linewidth]{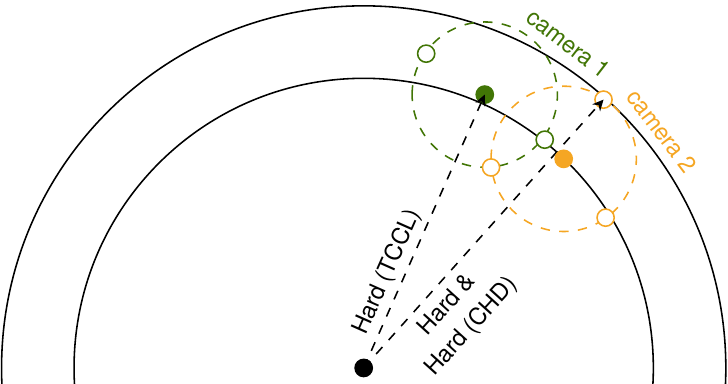}
        \caption{w/o noise}
        \label{fig:update_no_noise}
    \end{subfigure}
    \hfil
    \begin{subfigure}{0.47\linewidth}
        \includegraphics[width=\linewidth]{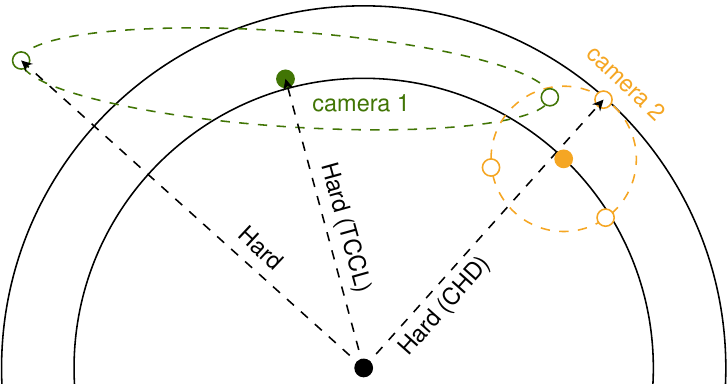}
        \caption{w/ noise}
        \label{fig:update_noise}
    \end{subfigure}
    \caption{Exemplar comparison among different memory hard sampling update strategies. Hollow points denote samples. Black solid points denote the cluster centroids in memory. Green and orange solid points denote the two camera centres, repectively. Black solid lines denote the equidistant lines. Black dashed arrows denote the directions of memory update.}
    \label{fig:hard_comparison}
\end{figure}

Another schematic diagram shown in Fig. \ref{fig:hard_comparison} illustrates the comparisons on the hard samples selected by \textbf{Hard(CHD)}, \textbf{Hard (TCCL)} and \textbf{Hard(CHD)}, respectively. Fig. \ref{fig:update_no_noise} constructs a scenario without noise, where the three update approaches capture similar hard sample and admissible direction to update memory.
Fig. \ref{fig:update_noise} constructs a scenario with noise, where both \textbf{Hard} and \textbf{Hard (TCCL)} select improper hard samples in the presence of noise (the one in the
upper left). However, \textbf{Hard (CHD)} is consistently able to recognise the hard sample that is far from the cluster centroid and closed to the camera centre, thus lead to a reasonable update direction.

\subsection{Summary of the Algorithm}
In the supervised Re-ID task \cite{JVCI19Spherereid}, a warm-up learning rate schedule can help set up a better initialisation status and thus result in a better performance, when a network is initialised by weights pre-trained on ImageNet.
Because the 3C framework is also initialised with the feature extractor pre-trained on ImageNet, 
a warm-up strategy is conducted in the beginning epochs for training the model. Such warm-up process not solely includes a fine-to-coarse then coarse-to-fine learning rate schedule as that in \cite{JVCI19Spherereid}, but also employs DBSCAN in the clustering stage, rather than HDC, to avoid the unexpected clustering confusion, due to the low discrimative power of the features learned in the early phase.
The learning procedure of the 3C framework is summarised in Algorithm \ref{alg:overall}.

\begin{algorithm}[ht]
\caption{\textbf{C}onfidence-Guided \textbf{C}lustering and \textbf{C}ontrastive Learning (3C)}
\label{alg:overall}
\begin{algorithmic}[1]
\REQUIRE ~~\\
$\mathcal{X}$, the unlabelled training data,\\
$f_{pre}$, the feature extractor pre-trained on ImageNet.\\
\ENSURE ~~\\
$f_{\theta}$, the trained feature extractor.

\STATE $f_{\theta}\leftarrow f_{pre}$;
\FOR{$epoch$ in $[1,num\_epochs]$}
    \STATE Extract features from $\mathcal{X}$ by $f_{\theta}$;
    \IF{$epoch <= num\_warmup$}
        \STATE Generate a pseudo labelled dataset $\mathbb{U}$ by DBSCAN;
    \ELSE
        \STATE Generate a pseudo labelled dataset $\mathbb{U}$ by HDC;
    \ENDIF
    \STATE Initial the cluster memory $\mathcal{M}$ by cluster centroids;
    \FOR{$iter$ in $[1,num\_iters]$}
        \STATE Sample $P \times K$ training images from $\mathbb{U}$;
        \STATE Calculate CIE of $P$ clusters with Eq. \eqref{eq:cam_info_entropy};
        \STATE Calculate weighted loss with Eq. \eqref{eq:weighted_loss};
        \STATE Update $f_{\theta}$ through back-propagation;
        \STATE Calculate CHD of $P \times K$ training images with Eq. \eqref{eq:hd_camera};
        \STATE Update $\mathcal{M}$ with Eqs. \eqref{eq:memory_update_hd_hard} and \eqref{eq:memory_update_hard};
    \ENDFOR
\ENDFOR
\RETURN $f_{\theta}$;
\end{algorithmic}
\end{algorithm}

\section{Experiments}
\label{sec:experiments}

\subsection{Datasets and Evaluation Protocol}
To evaluate the effectiveness of the proposed 3C framework, three unsupervised Re-ID benchmark datasets are used: Market-1501, MSMT17 and VeRi-776, where the first two are pedestrian datasets and the last one is a vehicle dataset.

\textbf{Market-1501} \cite{2015Market-1501} consists of 32,668 images of 1,501 pedestrians, which are captured by 6 cameras. The training dataset includes 12,936 images of 751 IDs, and the rest 19,732 images of another 750 IDs are used for testing. Moreover, 3,368 testing images are employed as query images and the rest testing images form a gallery set.

\textbf{MSMT17} \cite{2018MSMT17} is a large and challenging dataset, which includes 126,441 images of 4,101 pedestrians from 12 outdoor and 3 indoor cameras. The training dataset includes 32,621 images of 1,041 IDs. The rest 3,060 IDs contains  93,820 images,  which are split into a gallery set and a query set, containing 82,161 and 11,659 images, respectively.

\textbf{VeRi-776} \cite{2016VeRi-776} is a real-world vehicle Re-ID dataset, which comprises 51,038 images of 776 vehicles from 20 cameras. The training dataset consists of 37,781 images of 576 IDs. 
The rest 13,257 images of another 200 IDs are used for testing, where the gallery set contains 11,579 images, and the query set contains 1,678 images. 

\textbf{Evaluation Protocol}. In this paper, Cumulative Match Characteristic (CMC) \cite{2007CMC} scores and mean Average Precision (mAP) are employed to evaluate the performance of the methods. In particular, Rank-1, Rank-5 and Rank-10 scores are reported to represent the CMC curve. 
For fair comparisons, no post-processing technique (e.g. re-ranking \cite{2017Re-Ranking} and multi-query fusion \cite{2015Multi-query}) are adopted for evaluating the unsupervised Re-ID methods involved in this paper. The test query retrieval is carried out with the Euclidean distance.

\subsection{Implementation details}
\textbf{Backbone}. In this paper, Resnet50 \cite{2016Resnet} pre-trained on ImageNet \cite{2009ImageNet} is adopted as the backbone. In the light of the cluster contrast framework used in \cite{ACCV22ClusterContrast}, all sub-module layers after layer-4 are removed and a Generalised-Mean (GeM) pooling \cite{2018GeM} is added followed by the Batch Normalisation Neck (BNNeck) \cite{TM19BNN} and an $\ell_2$-normalisation layer to yield 2048-dimensional features.

\textbf{Data Augmentation}. The input images are resized to $256\times128$ for Market-1501 and MSMT17, and $224\times224$ for VeRi-776, respectively. Like most existing Re-ID methods, random horizontal flipping, padding with 10 pixels, random cropping and random erasing \cite{2020RandomErasing} are applied for data augmentation.

\textbf{Optimisation}. Adam \cite{2015Adam} is adopted as the optimiser with weight decay of $5\times10^{-4}$. The learning rate is initialised  as $3.5\times10^{-5}$ and increases to $3.5\times10^{-4}$ during a 10-epoch warm-up with DBSCAN, monotonically. Such value is kept during epochs 11-30. Then, the learning rate drops to $3.5\times10^{-5}$ during epochs 31-50, and further to $3.5\times10^{-6}$ during epochs 51-60.
Each mini-batch contains 256 images of 16 pseudo-labels and each pseudo-label contains 16 images.
The number of iterations per epoch is set to 200 for Market-1501, and 400 for MSMT17 and VeRi-776, respectively.

\textbf{Clustering}. For DBSCAN, the distance threshold between neighbours is set to 0.6. For $K$\!-means and HDC, $K$\!-means++ \cite{2007KMeans++} is utilised for initialisation. 
For DBSCAN and HDC, the Jaccard distance in \cite{2017Re-Ranking} is calculated as the sample-to-sample distance, where $k1$ and $k2$ are set to 30 and 6, respectively. For HDC, the Euler representation based cosine distance \cite{2013EulerRepresentation} is calculated as the sample-to-cluster distance.
For all clustering algorithms, the clusters have less that 4 samples are removed.

\textbf{Environment}. The proposed 3C framework is implemented based on PyTorch on Linux. All experiments are executed with a hardware environment with two Intel Xeon Gold 5218 CPU @2.30GHz and two NVIDIA RTX 3090 GPUs.

\subsection{Number of clusters for partition-based clustering}
\begin{figure}[!htb]
    \centering
    \begin{subfigure}{0.47\linewidth}
        \includegraphics[width=\linewidth]{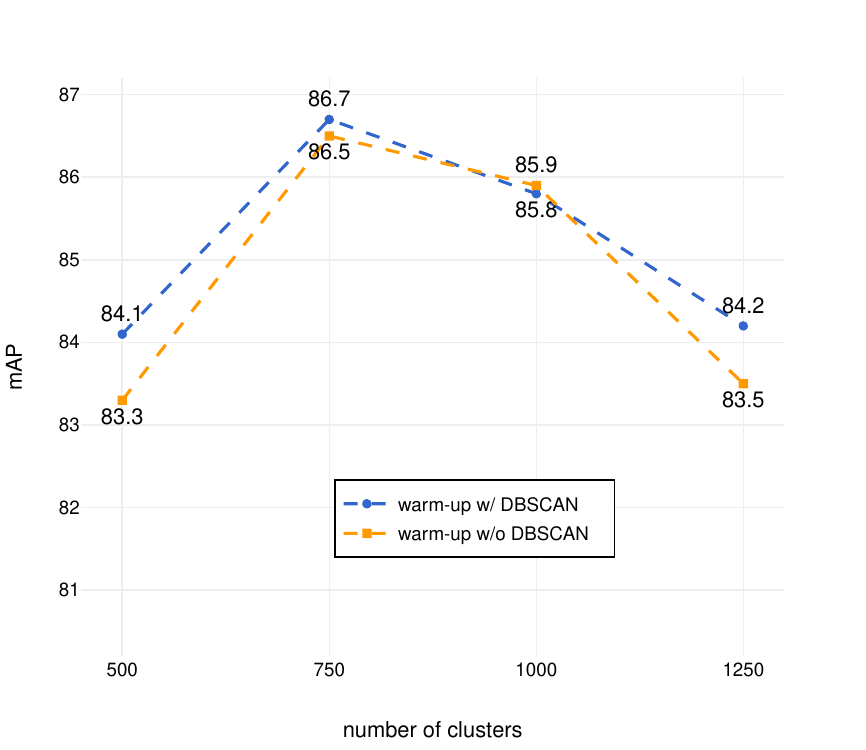}
        \caption{Market-1501 (HDC)}
    \end{subfigure}
    \hfil
    \begin{subfigure}{0.47\linewidth}
        \includegraphics[width=\linewidth]{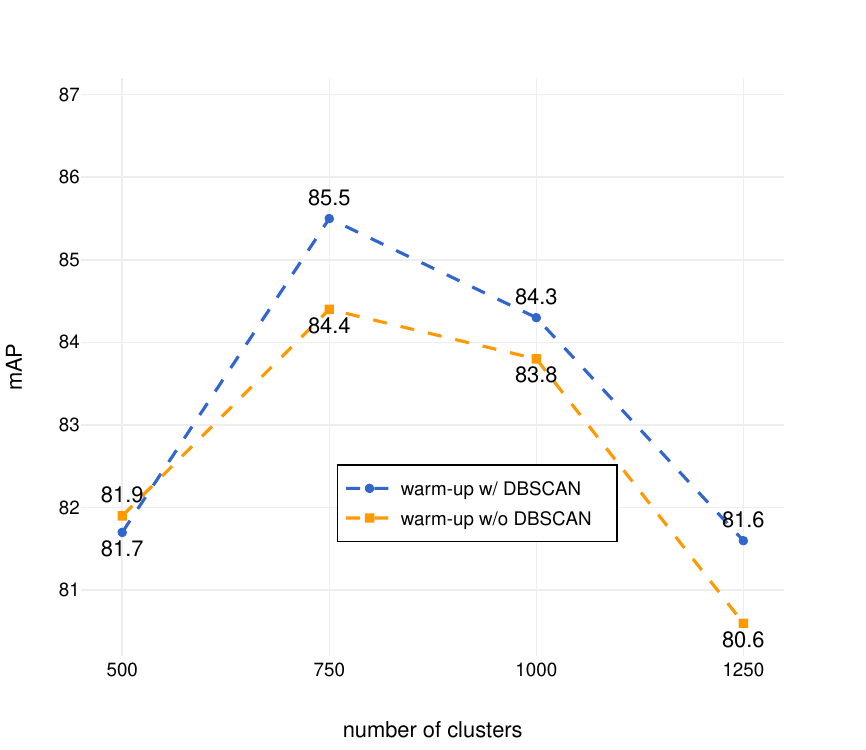}
        \caption{Market-1501 ($K$\!-means)}
    \end{subfigure}
    
    \begin{subfigure}{0.47\linewidth}
        \includegraphics[width=\linewidth]{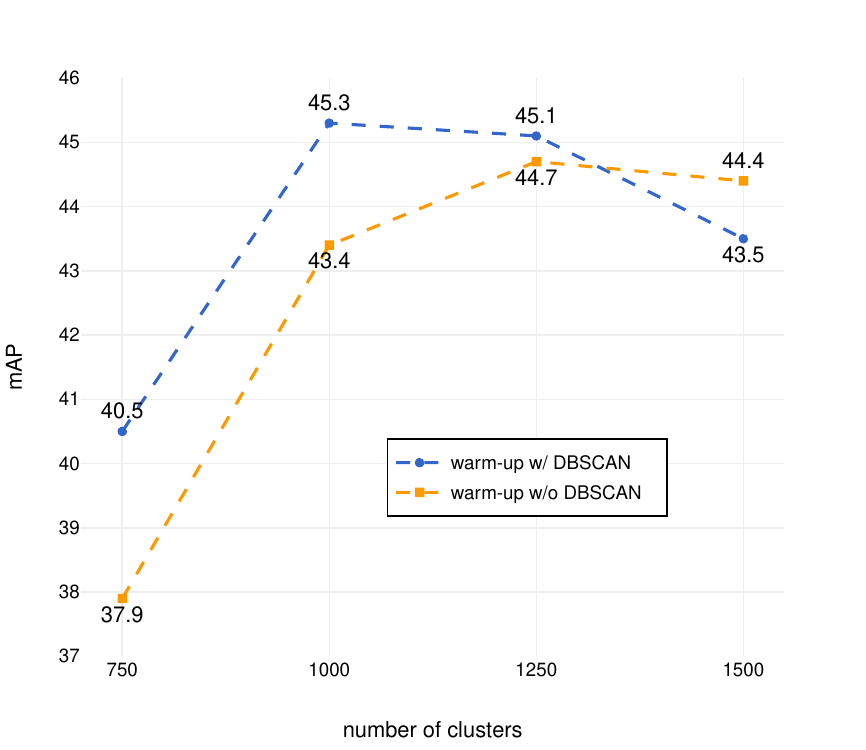}
        \caption{MSMT17 (HDC)}
    \end{subfigure}
    \hfil
    \begin{subfigure}{0.47\linewidth}
        \includegraphics[width=\linewidth]{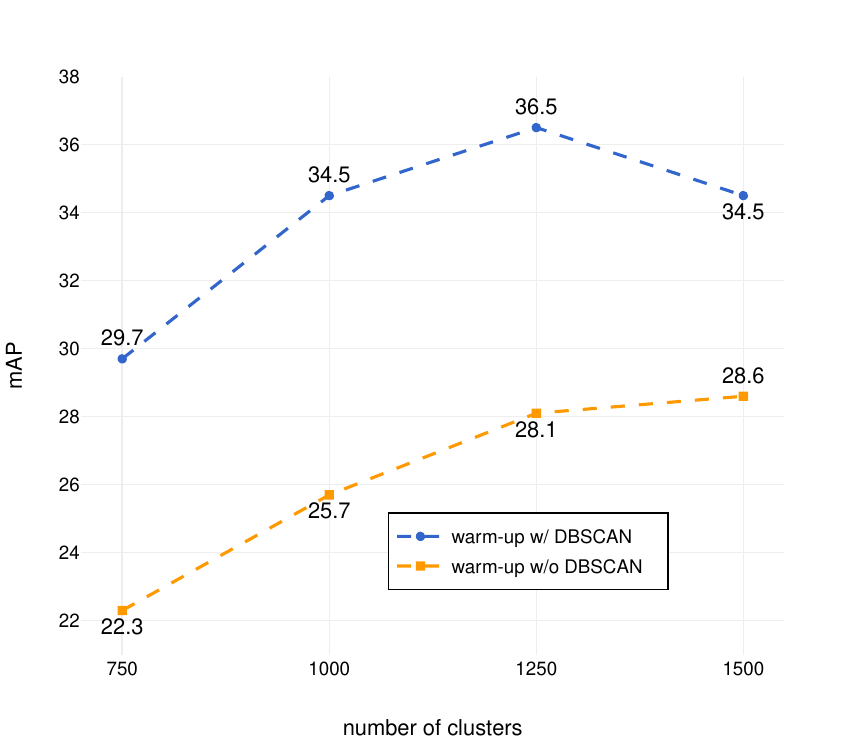}
        \caption{MSMT17 ($K$\!-means)}
    \end{subfigure}
    
    \begin{subfigure}{0.47\linewidth}
        \includegraphics[width=\linewidth]{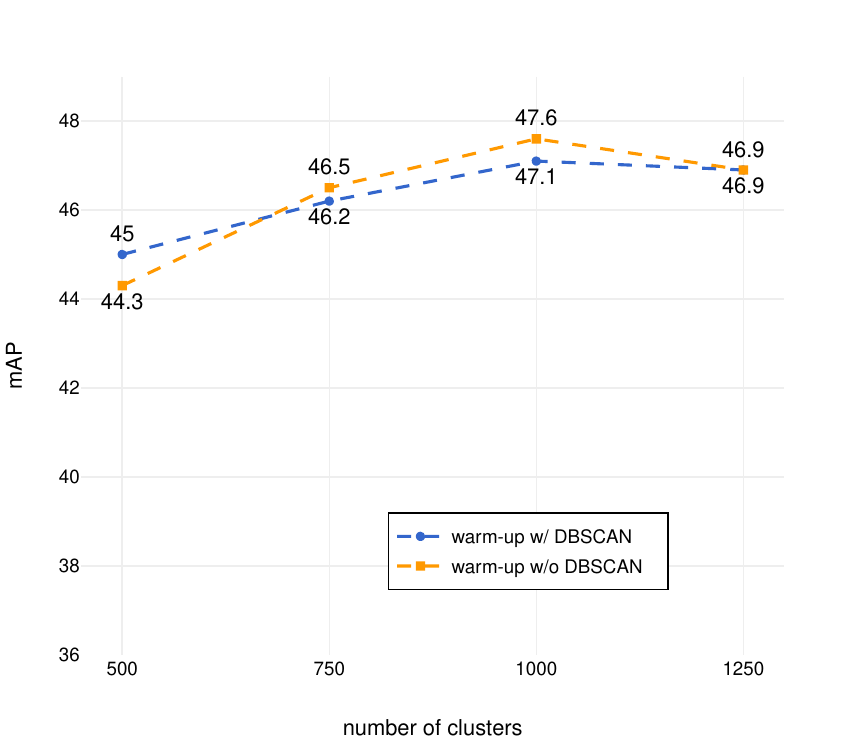}
        \caption{VeRi-776 (HDC)}
        \label{fig:veri_HDC}
    \end{subfigure}
    \hfil
    \begin{subfigure}{0.47\linewidth}
        \includegraphics[width=\linewidth]{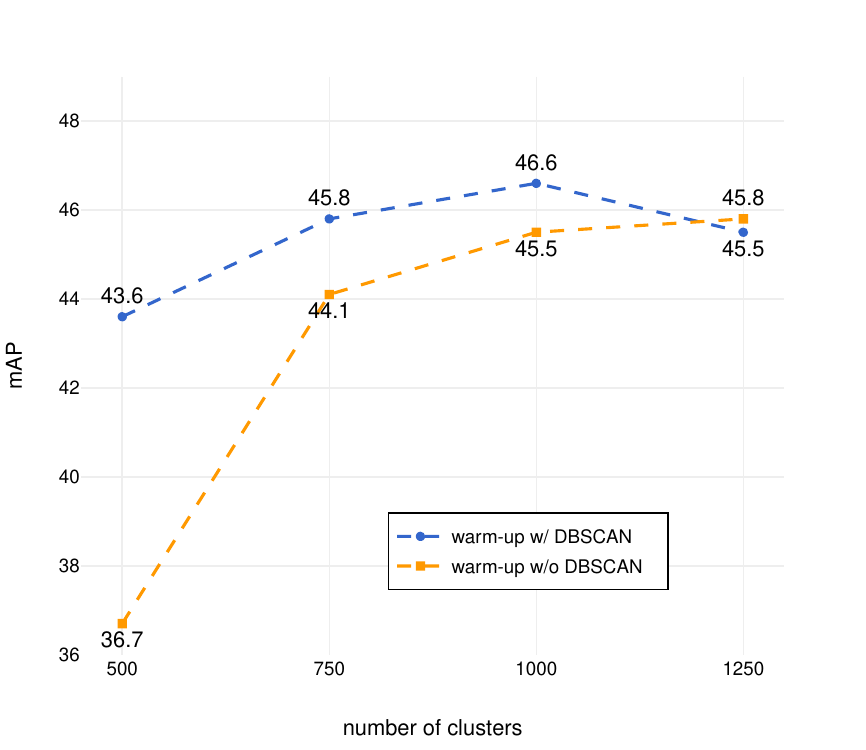}
        \caption{VeRi-776 ($K$\!-means)}
        \label{fig:veri_kmeans}
    \end{subfigure}
    \caption{Evaluation on different number of clusters.}
    \label{fig:paramater_num_clusters}
\end{figure}

Fig. \ref{fig:paramater_num_clusters} illustrates the results of  HDC and $K$\!-means with different numbers of clusters on the three used datasets.

In order to determine the respective numbers of clusters of partition-based clustering algorithms on the three used datasets, the results of the 3C framework, implemented via HDC or $K$\!-means with different numbers of clusters, are illustrated in Fig. \ref{fig:paramater_num_clusters}. In particular, each study is conducted in two cases: with or without the use of DBSCAN for warm-up.

\begin{table*}[]
\caption{Ablation studies on Clustering, CID and warm-up with DBSCAN.}
\label{tb:ablation1}
\centering
\begin{tabular}{ccc|cccc|cccc|cccc}
\hline
\multirow{2}{*}{Clustering}  & \multirow{2}{*}{CIE} & \multirow{2}{*}{\begin{tabular}[c]{@{}c@{}}Warm-up\\w/ DBSCAN\end{tabular}} & \multicolumn{4}{c|}{Market-1501} & \multicolumn{4}{c|}{MSMT17} & \multicolumn{4}{c}{VeRi-776} \\ \cline{4-15} 
 &  &   & mAP & R1 & R5 & R10 & mAP & R1 & R5 & R10 & mAP & R1 & R5 & R10 \\ \hline
 \multirow{2}{*}{DBSCAN} &  &  & 84.6 & 92.9 & 96.8 & 97.5 & 21.9 & 45.8 & 56.7 & 61.6 & 38.8 & 80.3 & 83.4 & 86.2 \\
 & \checkmark &  & 84.6 & 93.6 & 97.3 & 98.2 & \textbf{47.4} & \textbf{73.7} & \textbf{83.5} & \textbf{86.5} & 44.5 & 86.2 & 89.8 & 91.7 \\ \hline
 \multirow{3}{*}{$K$\!-means} &  &  & 82.2 & 92.7 & 96.8 & 97.9 & 18.9 & 43.7 & 54.8 & 59.7 & 27.4 & 62.9 & 67.8 & 72.2 \\
 & \checkmark &  & 84.4 & 93.6 & 96.9 & 97.8 & 25.7 & 54.1 & 65.8 & 70.6 & 45.5 & 88.2 & 92.0 & 93.5 \\
 & \checkmark & \checkmark & 85.5 & 94.0 & 97.3 & 98.1 & 34.5 & 65.2 & 75.4 & 79.4 & 46.6 & 89.8 & 92.8 & 94.3 \\ \hline
 \multirow{3}{*}{HDC}  &  &  & 85.4 & 94.1 & 97.5 & 98.2 & 34.0 & 62.9 & 74.1 & 78.2 & 38.8 & 80.3 & 83.6 & 85.3 \\
  & \checkmark &  & 86.5 & 94.3 & 97.7 & 98.2 & 43.4 & 72.4 & 82.0 & 84.8 & \textbf{47.6} & 90.1 & \textbf{93.5} & \textbf{95.1} \\
 & \checkmark & \checkmark & \textbf{86.7} & \textbf{94.7} & \textbf{97.8} & \textbf{98.6} & 45.3 & 73.1 & 82.7 & 85.5 & 47.1 & \textbf{90.6} & 93.2 & \textbf{95.1} \\ \hline
\end{tabular}
\end{table*}

\begin{table*}[]
\caption{Ablation studies on different memory update strategies with/without CIE.}
\label{tb:ablation2}
\centering
\begin{tabular}{lc|cccc|cccc|cccc}
\hline
\multirow{2}{*}{Update Strategy} & \multirow{2}{*}{CIE} & \multicolumn{4}{c|}{Market-1501} & \multicolumn{4}{c|}{MSMT17} & \multicolumn{4}{c}{VeRi-776} \\ \cline{3-14} 
 &  & mAP & R1 & R5 & R10 & mAP & R1 & R5 & R10 & mAP & R1 & R5 & R10 \\ \hline
Vanilla & \multirow{4}{*}{} & 84.7 & 93.1 & 97.2 & 98.0 & 32.2 & 62.0 & 73.2 & 77.2 & 39.0 & 82.2 & 86.4 & 89.4 \\
Hard &  & 85.2 & 93.5 & 97.1 & 97.8 & 29.6 & 56.8 & 68.6 & 73.2 & 38.8 & 80.3 & 83.6 & 85.3 \\
Hard (TCCL) &  & 84.0 & 93.4 & 96.8 & 97.5 & 39.0 & 68.6 & 79.1 & 82.8 & 42.3 & 86.5 & 90.9 & 93.0 \\
Hard (CHD) &  & 86.5 & 93.8 & 97.6 & 98.3 & 32.7 & 60.2 & 71.1 & 75.6 & 39.1 & 81.1 & 84.2 & 86.4 \\ \hline
Vanilla & \multirow{4}{*}{\checkmark} & 85.4 & 93.4 & 97.0 & 98.0 & 41.6 & 70.3 & 80.5 & 84.0 & 45.5 & 89.0 & 92.9 & 94.5 \\
Hard &  & 86.5 & 94.5 & 97.6 & 98.4 & 43.0 & 70.7 & 80.7 & 83.7 & 46.7 & 89.8 & 93.6 & \textbf{95.1} \\
Hard (TCCL) &  & 84.5 & 93.8 & 97.7 & 98.4 & 40.8 & 69.8 & 80.3 & 83.8 & 46.4 & 89.8 & \textbf{93.7} & 94.8 \\
Hard (CHD) &  & \textbf{86.7} & \textbf{94.7} & \textbf{97.8} & \textbf{98.6} & \textbf{45.3} & \textbf{73.1} & \textbf{82.7} & \textbf{85.5} & \textbf{47.1} & \textbf{90.6} & 93.2 & \textbf{95.1} \\ \hline
\end{tabular}
\end{table*}

For Market-1501, both HDC and $K$\!-means approximately reach the best results when the numbers of clusters (750) are close to that of real IDs (751). But on MSMT17, such consistency (1000 clusters v.s. 1041 IDs) is only observed in the case of HDC with DBSCAN for warm-up. For HDC without DBSCAN for warm-up, the fail may be led by the low discriminative nature of features learned in the early stages. And the inferior results of $K$\!-means is mainly due to its sensitivity to noise in the complex scenario.

Different from pedestrian datasets, in VeRi-776, different camera angles will lead to huge shape variation of images that belong to the same vehicle. 
Due to such drastic internal diversity of individual vehicles, the cluster corresponding to each ID for the vehicle Re-ID task is better to be further divided into several sub-clusters. As shown in  Figs. \ref{fig:veri_HDC} and \ref{fig:veri_kmeans}, HDC and $K$\!-means get the better results with 1000 clusters, compared to those with 500 clusters (around the 576 real IDs).  

In summary, the numbers of clusters in the subsequent experiments are set to 750, 1,000 and 1,000 for Market-1501, MSMT17 and VeRi-776, respectively.

\subsection{Ablation studies}
To confirm the validity of the three confidence metrics and the DBSCAN warm-up approach used in the proposed 3C framework, ablation studies are conducted by decomposing the 3C framework into four modules: clustering, (the use of) CIE, warm-up with DBSCAN and (the use of) CHD. The studies on the first three components are shown in Table \ref{tb:ablation1}. And in Table \ref{tb:ablation2}, a detailed comparison of different memory update strategies is presented with or without the use of CIE.

\subsubsection{Effectiveness of HDC}
\label{sec:effectiveness_HDC}
As shown in Table \ref{tb:ablation1}, compared to DBSCAN, HDC surges ahead on Market-1501 and VeRi-776 but lags behind on MSMT17. As shown in Fig. \ref{fig:CIE_illustration}, images of some IDs (such as ID 0936) span indoors and outdoors on MSMT17. Such drastic changes in the scene will pose partition difficulties for HDC. 
However,  DBSCAN can split an real ID (cluster) into multiple sub-IDs (sub-clusters) to fit the complex scenes in MSMT17. 

Compared to $K$\!-means, HDC can be regarded as a superior alternative in terms of experimental results and theory.
It is noteworthy that, with all the components, even the under-appreciated $K$\!-means method outperforms the favoured DBSCAN on Market-1501 and VeRi-776. This implies the potential of partition-based clustering algorithms for the unsupervised Re-ID tasks.

\subsubsection{Effectiveness of CIE}
\label{sec:effectiveness_CIE}
CIE can As demonstrated by the results in Tables \ref{tb:ablation1}, the use of CIE improves results in most cases, especially for the two complex large-scale datasets. In particular, on MSMT17, the respective results of mAP/Rank-1 accuracy of DBSCAN, $K$\!-means and HDC improve 25.5\%/27.9\%, 6.8\%/10.4\% and 9.4\%/9.5\%. On VeRi-776, the improvements are 5.7\%/5.9\%, 18.1\%/25.3\% and 9.3\%/9.6\%, respectively.

\begin{figure}[!htb]
    \centering
    \begin{subfigure}{0.47\linewidth}
        \includegraphics[width=\linewidth]{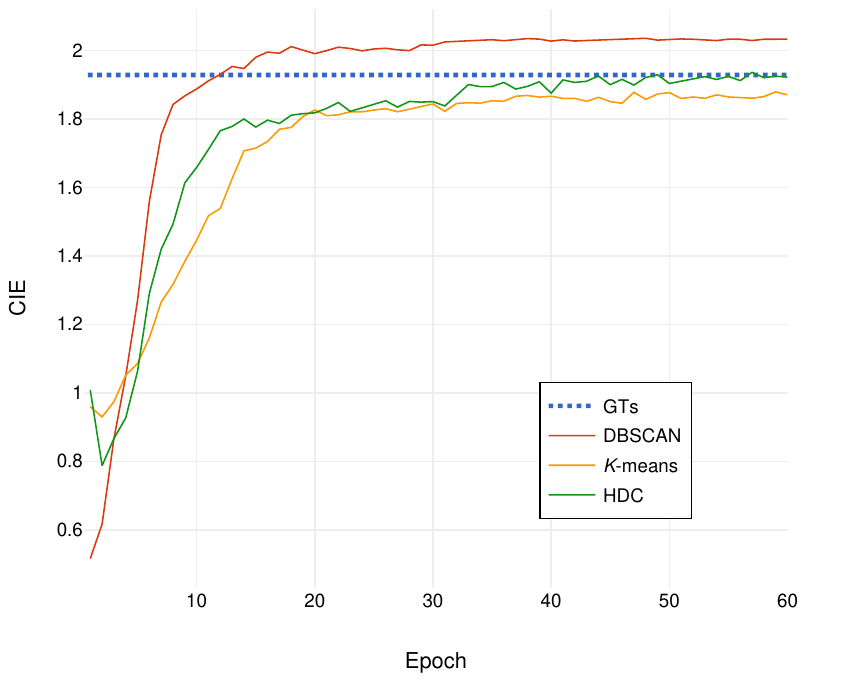}
        \caption{Market-1501}
    \end{subfigure}
    \hfil
    \begin{subfigure}{0.47\linewidth}
        \includegraphics[width=\linewidth]{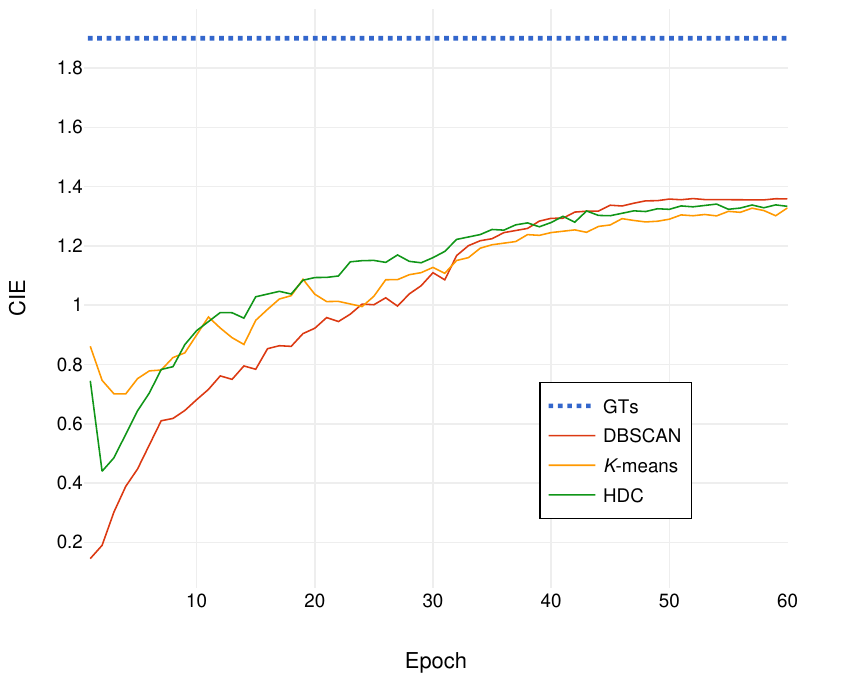}
        \caption{MSMT17}
    \end{subfigure}
    
    \begin{subfigure}{0.47\linewidth}
        \includegraphics[width=\linewidth]{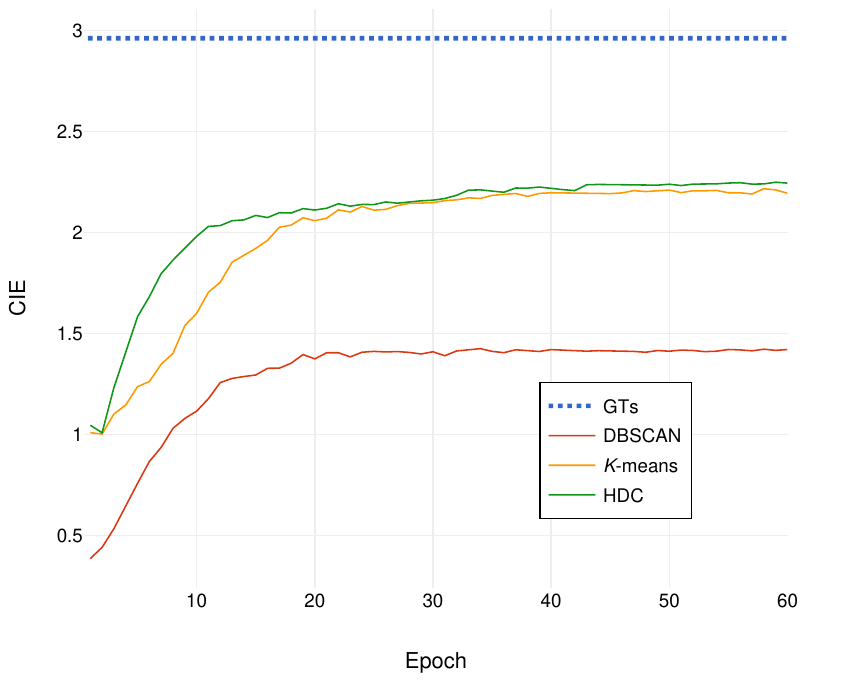}
        \caption{VeRi-776}
    \end{subfigure}
    \hfil
    \begin{subfigure}{0.47\linewidth}
        \includegraphics[width=\linewidth]{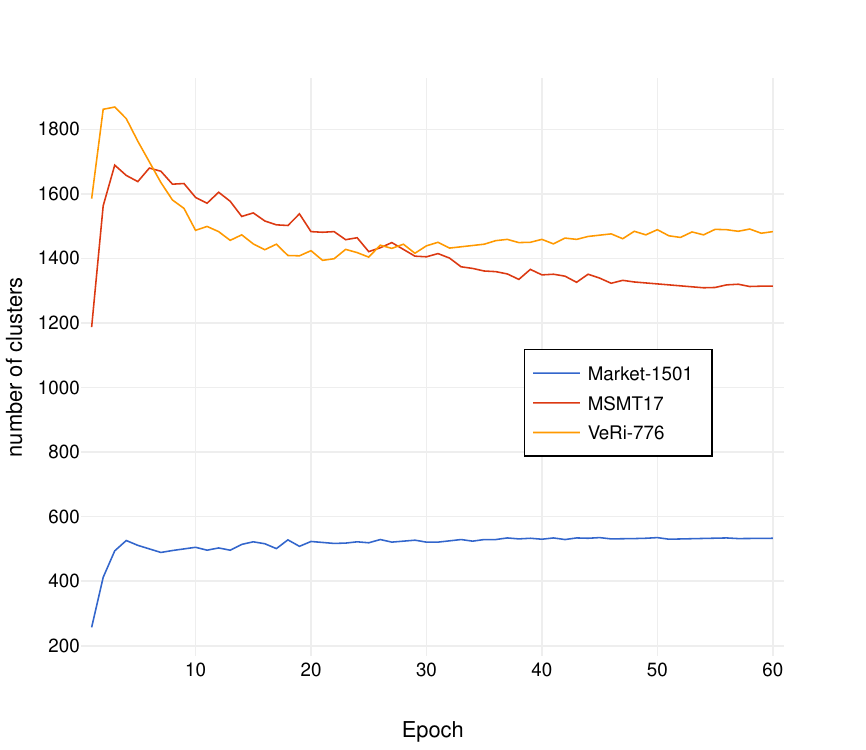}
        \caption{number of clusters (DBSCAN)}
    \end{subfigure}
    \caption{CIE curves and cluster number curves of DBSCAN.}
    \label{fig:CIE_curves}
\end{figure}

To further analyse the role of CIE in training the 3C framework, the respect curves of the average  CIE of all clusters by DBSCAN, $K$\!-means and HDC, with respect to learning epochs are shown in Fig. \ref{fig:CIE_curves}. 
The ground truths (GTs) are the average CIE values of all real IDs in all datasets. 
In addition, to have insight into the results of the DBSCAN-based variant of the 3C framework, the numbers of clusters by DBSCAN are also illustrated in Fig. \ref{fig:CIE_curves}.

The results show that in general, the average CIE values of all three clustering methods grow towards the GTs as the learning progresses. On Market-1501, DBSCAN produces 560 clusters, which is less than 751 real IDs and 750 clusters of HDC and $K$\!-means. Thus, the CIE curve of DBSCAN is over the GT. On MSMT17, DBSCAN converges to 1300 clusters, which is more than 1000 clusters of HDC and $K$\!-means. However, the average CIE values of DBSCAN, HDC and $K$\!-means are almost identical at the end of the training stage. This implies that DBSCAN gets a large amount of CIE than those of  HDC and $K$\!-means. As for VeRi-776, DBSCAN has a rather lower average of CIE than those of HDC and $K$\!-means. But it still generates more clusters (1400 v.s. 1000). This indicates that the quantity of CIE by DBSCAN is not much larger, or even lower, than those of HDC and $K$\!-means. 

As analysed previously, CIE may be used as an indicator to forecast the performance of a Re-ID model, because the clusters with large CIE are informative to training the model. It is interesting that such inference is consistent with the experimental results. Due to the higher values of CIE, the DBSCAN variant of the 3C framework achieves better results on MSMT17, and the 3C framework with HDC performs better on  Market-1501 and VeRi-776.

\subsubsection{Effectiveness of warm-up with DBSCAN}
\label{sec:warm-up}
The proposed 3C framework can use a warm-up process with DBSCAN to boost the model with the low discriminative power of the early features. The comparative results of $K$\!-means and HDC in Table \ref{tb:ablation1} demonstrate that the warm-up process with DBSCAN
can help partition-based clustering algorithms, especially $K$\!-means, generate decent pseudo-labels, in terms of quality and quantity, to guidance the following contrastive learning.

\subsubsection{Effectiveness of CHD}
\label{sec:hd_hard}
As shown in Table \ref{tb:ablation2}, \textbf{Hard (CHD)} obtains the best results in 11 cases out of 12, in presence of CIE. 
Specifically, when CIE is absent, the camera information is not sufficient for \textbf{Hard (CHD)} in each cluster. Therefore, \textbf{Hard (CHD)} performs worse than \textbf{Hard (TCCL)} on MSMT17 and VeRi-776.
However, with the use of CIE, \textbf{Hard (CHD)} can identify the hard sample with valuable camera information. In this case, \textbf{Hard (CHD)} takes the superiority in updating memory.

\begin{figure}[!htb]
    \centering
    \begin{subfigure}{0.47\linewidth}
        \includegraphics[width=\linewidth]{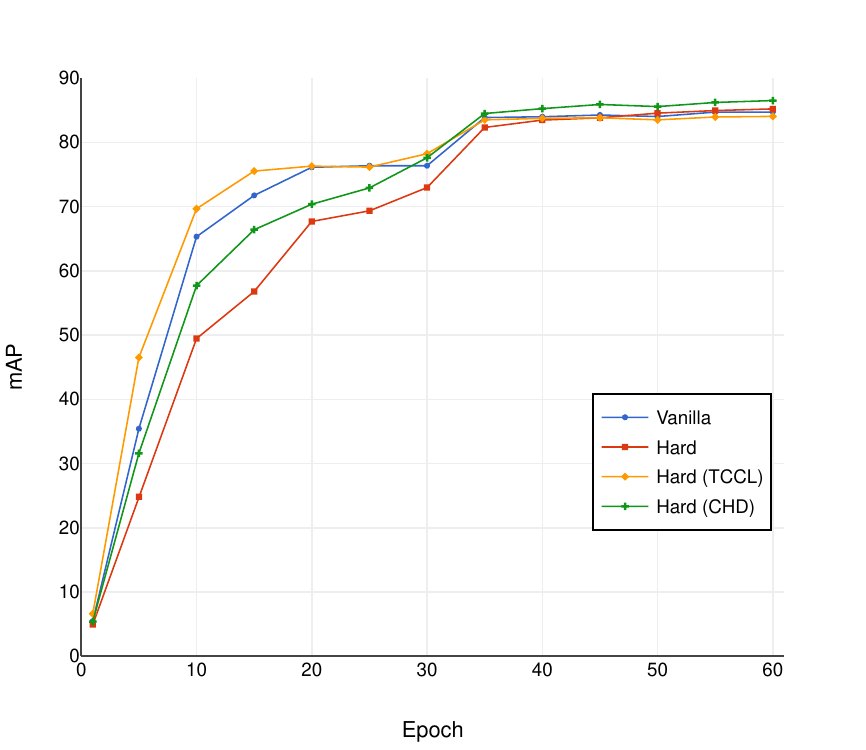}
        \caption{Market-1501 (w/o CIE)}
    \end{subfigure}
    \hfil
    \begin{subfigure}{0.47\linewidth}
        \includegraphics[width=\linewidth]{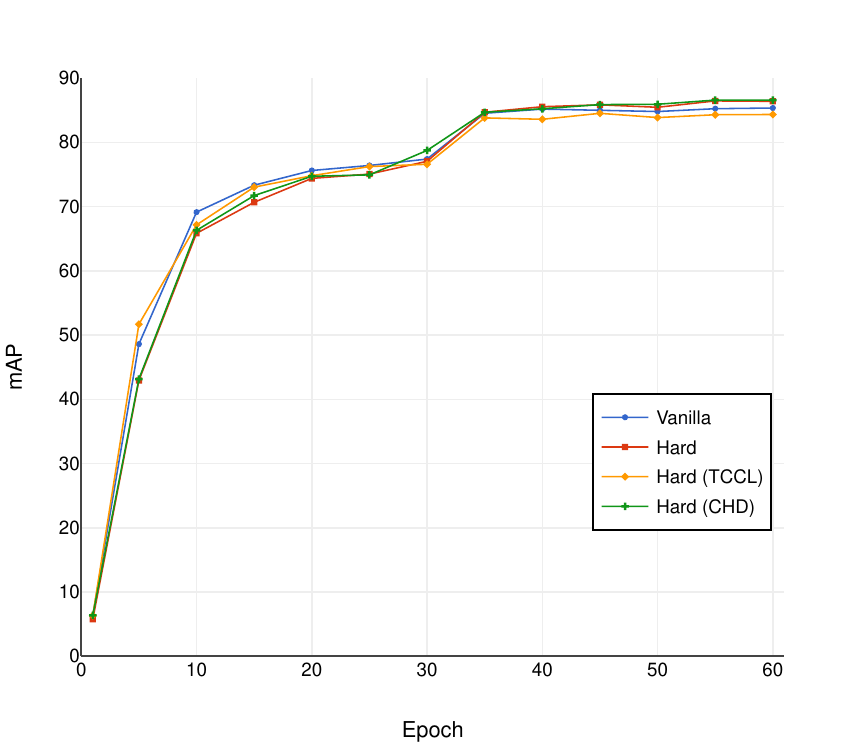}
        \caption{Market-1501 (w/ CIE)}
    \end{subfigure}

    \begin{subfigure}{0.47\linewidth}
        \includegraphics[width=\linewidth]{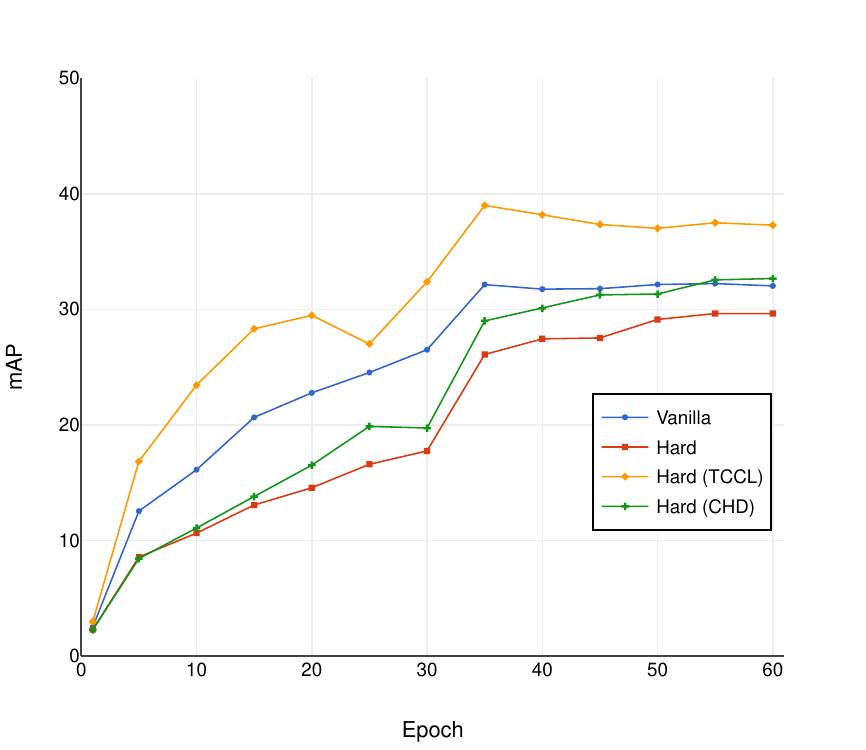}
        \caption{MSMT17 (w/o CIE)}
    \end{subfigure}
    \hfil
    \begin{subfigure}{0.47\linewidth}
        \includegraphics[width=\linewidth]{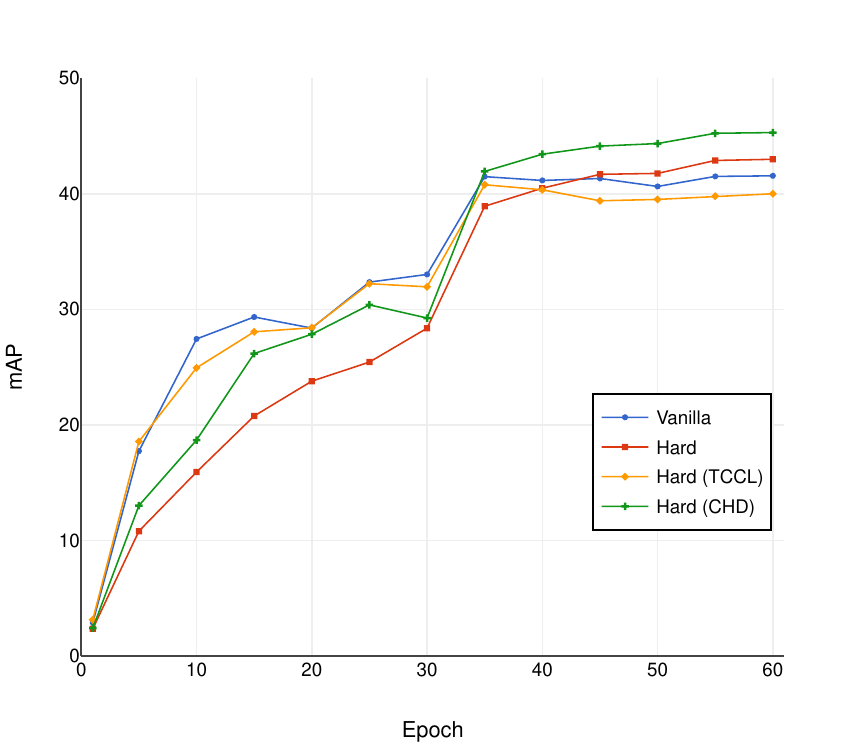}
        \caption{MSMT17 (w/ CIE)}
    \end{subfigure}

    \begin{subfigure}{0.47\linewidth}
        \includegraphics[width=\linewidth]{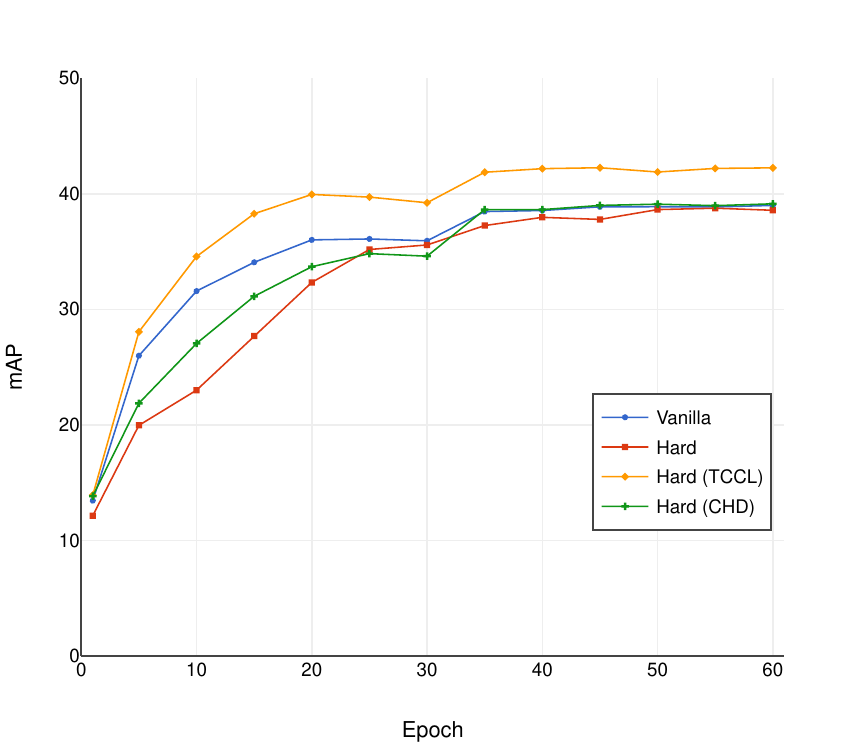}
        \caption{VeRi-776 (w/o CIE)}
    \end{subfigure}
    \hfil
    \begin{subfigure}{0.47\linewidth}
        \includegraphics[width=\linewidth]{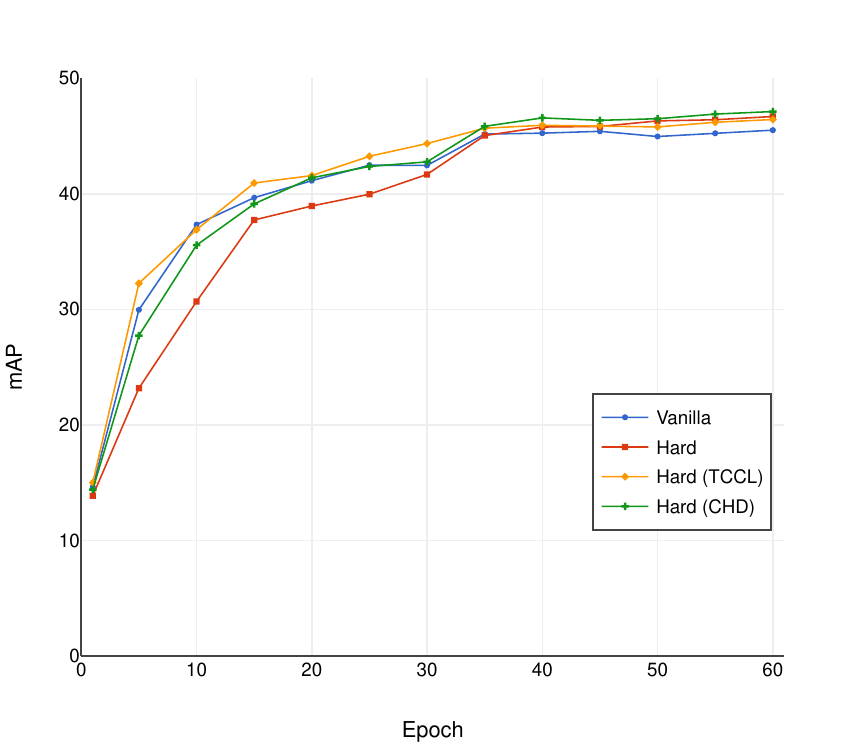}
        \caption{VeRi-776 (w/ CIE)}
    \end{subfigure}
    \caption{Evaluation on different memory update strategies.}
    \label{fig:updating_strategy_line}
\end{figure}

The leading role of the combined action of CHD and CIE is also illustrated in Fig. \ref{fig:updating_strategy_line}. 
It can be seen that, when CIE is not used, \textbf{Hard (TCCL)} can more quickly boost the network performance in the early training stages, compared to the others. But aided by the camera information provides by CIE, 
\textbf{Hard (CHD)} can orient the memory update to a hard (for cluster) and promising (for camera) direction.

\subsection{Comparison with the state-of-the-art methods}

\begin{table*}[]
\caption{Comparison with state-of-art methods on Market-1501, MSMT17 and VeRi-776.}
\label{tb:sota}
\centering
\begin{tabular}{l|c|cccc|cccc|cccc}
\hline
\multicolumn{1}{c|}{\multirow{2}{*}{Method}} & \multicolumn{1}{c|}{\multirow{2}{*}{Reference}} & \multicolumn{4}{c|}{Market-1501} & \multicolumn{4}{c|}{MSMT17} & \multicolumn{4}{c}{VeRi-776} \\ \cline{3-14} 
\multicolumn{1}{c|}{} & \multicolumn{1}{c|}{} & mAP & R1 & R5 & \multicolumn{1}{c|}{R10} & mAP & R1 & R5 & \multicolumn{1}{c|}{R10} & mAP & R1 & R5 & R10 \\ \hline
\textbf{Fully Unsupervised} & \multicolumn{1}{l|}{} & \multicolumn{1}{l}{} & \multicolumn{1}{l}{} & \multicolumn{1}{l}{} & \multicolumn{1}{l|}{} & \multicolumn{1}{l}{} & \multicolumn{1}{l}{} & \multicolumn{1}{l}{} & \multicolumn{1}{l|}{} & \multicolumn{1}{l}{} & \multicolumn{1}{l}{} & \multicolumn{1}{l}{} & \multicolumn{1}{l}{} \\
\multicolumn{1}{l|}{BUC \cite{AAAI19BUC}} & \multicolumn{1}{c|}{AAAI'19} & 38.3 & 66.2 & 79.6 & \multicolumn{1}{c|}{84.5} & - & - & - & \multicolumn{1}{c|}{-} & - & - & - & - \\
\multicolumn{1}{l|}{HCT \cite{CVPR20HCT}} & \multicolumn{1}{c|}{CVPR'20} & 56.4 & 80.0 & 91.6 & \multicolumn{1}{c|}{95.2} & - & - & - & \multicolumn{1}{c|}{-} & - & - & - & - \\
\multicolumn{1}{l|}{SpCL \cite{NPIS20SpCL}} & \multicolumn{1}{c|}{NeurIPS'20} & 73.1 & 88.1 & 95.1 & \multicolumn{1}{c|}{97.0} & 19.1 & 42.3 & 55.6 & \multicolumn{1}{c|}{61.2} & 36.9 & 79.9 & 86.8 & 89.9 \\
\multicolumn{1}{l|}{ICE \cite{ICCV21ICE}} & \multicolumn{1}{c|}{ICCV'21} & 79.5 & 92.0 & 97.0 & \multicolumn{1}{c|}{98.1} & 29.8 & 59.0 & 71.7 & \multicolumn{1}{c|}{77.0} & - & - & - & - \\
\multicolumn{1}{l|}{CACL \cite{TIP22CACL}} & \multicolumn{1}{c|}{TIP'22} & 80.9 & 92.7 & 97.4 & \multicolumn{1}{c|}{98.5} & 23.0 & 48.9 & 61.2 & \multicolumn{1}{c|}{66.4} & - & - & - & - \\
\multicolumn{1}{l|}{Cluster-Contrast\cite{ACCV22ClusterContrast}} & \multicolumn{1}{c|}{ACCV'22} & 83.0 & 92.9 & 97.2 & \multicolumn{1}{c|}{98.0} & 33.0 & 62.0 & 71.8 & \multicolumn{1}{c|}{76.7} & 40.8 & 86.2 & 90.5 & 92.8 \\
\multicolumn{1}{l|}{ISE \cite{CVPR22ISE}} & \multicolumn{1}{c|}{CVPR'22} & 85.3 & 94.3 & \textbf{98.0} & \multicolumn{1}{c|}{\ul 98.8} & 37.0 & 67.6 & 77.5 & \multicolumn{1}{c|}{81.0} & - & - & - & - \\
\multicolumn{1}{l|}{RTMem \cite{TIP23RTMem}} & \multicolumn{1}{c|}{TIP'23} & {\ul 86.5} & 94.3 & {\ul 97.9} & \multicolumn{1}{c|}{98.5} & 38.5 & 63.3 & 75.4 & \multicolumn{1}{c|}{79.6} & 44.2 & 85.2 & 89.6 & 92.0 \\
\multicolumn{1}{l|}{LP\cite{TIP23LP}} & \multicolumn{1}{c|}{TIP'23} & 85.8 & {\ul 94.5} & 97.8 & \multicolumn{1}{c|}{98.7} & 39.5 & 67.9 & 78.0 & \multicolumn{1}{c|}{81.6} & - & - & - & - \\
\multicolumn{1}{l|}{AdaMG \cite{TCSVT23AdaMG}} & \multicolumn{1}{c|}{TCSVT'23} & 84.6 & 93.9 & {\ul 97.9} & \multicolumn{1}{c|}{\textbf{98.9}} & 38.0 & 66.3 & 76.9 & \multicolumn{1}{c|}{80.6} & 41.0 & 86.2 & 90.6 & 93.1 \\
\multicolumn{1}{l|}{DCCT \cite{TCSVT23DCCT}} & \multicolumn{1}{c|}{TCSVT'23} & 86.3 & 94.4 & 97.7 & \multicolumn{1}{c|}{98.5} & 41.8 & 68.7 & 79.0 & \multicolumn{1}{c|}{82.6} & - & - & - & - \\
\multicolumn{1}{l|}{DCMIP \cite{ICCV23DCMIP}} & \multicolumn{1}{c|}{ICCV'23} & \textbf{86.7} & \textbf{94.7} & \textbf{98.0} & \multicolumn{1}{c|}{{\ul 98.8}} & 40.9 & 69.3 & 79.7 & \multicolumn{1}{c|}{83.6} & - & - & - & - \\
\multicolumn{1}{l|}{DHCCN \cite{TCSVT24DHCCN}} & \multicolumn{1}{c|}{TCSVT'24} & 85.6 & 94.1 & - & \multicolumn{1}{c|}{-} & 36.4 & 65.9 & - & \multicolumn{1}{c|}{-} & - & - & - & - \\ \hline
\textbf{Camera aware} & \multicolumn{1}{l|}{} & \multicolumn{1}{l}{} & \multicolumn{1}{l}{} & \multicolumn{1}{l}{} & \multicolumn{1}{l|}{} & \multicolumn{1}{l}{} & \multicolumn{1}{l}{} & \multicolumn{1}{l}{} & \multicolumn{1}{l|}{} & \multicolumn{1}{l}{} & \multicolumn{1}{l}{} & \multicolumn{1}{l}{} & \multicolumn{1}{l}{} \\
\multicolumn{1}{l|}{SSL \cite{CVPR20SSL}} & \multicolumn{1}{c|}{CVPR'20} & 37.8 & 71.7 & 83.8 & \multicolumn{1}{c|}{87.4} & - & - & - & \multicolumn{1}{c|}{-} & - & - & - & - \\
\multicolumn{1}{l|}{IICS \cite{CVPR21IICS}} & \multicolumn{1}{c|}{CVPR'21} & 72.9 & 89.5 & 95.2 & \multicolumn{1}{c|}{97.0} & 26.9 & 56.4 & 68.8 & \multicolumn{1}{c|}{73.4} & - & - & - & - \\
\multicolumn{1}{l|}{CAP\cite{AAAI21CAP}} & \multicolumn{1}{c|}{AAAI'21} & 79.2 & 91.4 & 96.3 & \multicolumn{1}{c|}{97.7} & 36.9 & 67.4 & 78.0 & \multicolumn{1}{c|}{81.4} & - & - & - & - \\
\multicolumn{1}{l|}{ICE \cite{ICCV21ICE}} & \multicolumn{1}{c|}{ICCV'21} & 82.3 & 93.8 & 97.6 & \multicolumn{1}{c|}{98.4} & 38.9 & 70.2 & 80.5 & \multicolumn{1}{c|}{84.4} & - & - & - & - \\
\multicolumn{1}{l|}{PPLR \cite{CVPR22PPLR}} & \multicolumn{1}{c|}{CVPR'22} & 84.4 & 94.3 & 97.8 & \multicolumn{1}{c|}{98.6} & 42.2 & {\ul 73.3} & \textbf{83.5} & \multicolumn{1}{c|}{\textbf{86.5}} & 43.5 &  88.3 & 92.7 & {\ul 94.4} \\
\multicolumn{1}{l|}{CCL \cite{TCSVT23CCL}} & \multicolumn{1}{c|}{TCSVT'23} & 85.3 & 94.1 & 97.8 & \multicolumn{1}{c|}{{\ul 98.8}} & 41.8 & 71.4 & 81.8 & \multicolumn{1}{c|}{85.1} & 42.6 & 87.0 & - & - \\
\multicolumn{1}{l|}{HCACE \cite{TMM24HCACE}} & \multicolumn{1}{c|}{TMM'24} & 83.4 & 93.7 & 97.5 & \multicolumn{1}{c|}{98.1} & 41.6 & 72.4 & 81.8 & \multicolumn{1}{c|}{84.9} & - & - & - & - \\
\multicolumn{1}{l|}{3C (DBSCAN)} & \multicolumn{1}{c|}{This Paper} & 84.6 & 93.6 & 97.3 & \multicolumn{1}{c|}{98.2} & \textbf{47.4} & \textbf{73.7} & \textbf{83.5} & \multicolumn{1}{c|}{\textbf{86.5}} & 44.5 & 86.2 & 89.8 & 91.7 \\
\multicolumn{1}{l|}{3C ($K$\!-means)} & \multicolumn{1}{c|}{This Paper} & 85.5 & 94.0 & 97.3 & \multicolumn{1}{c|}{98.1} & 34.5 & 65.2 & 75.4 & \multicolumn{1}{c|}{79.4} & {\ul 46.6} & {\ul 89.8} & {\ul 92.8} & 94.3 \\
\multicolumn{1}{l|}{3C} & \multicolumn{1}{c|}{This Paper} & \textbf{86.7} & \textbf{94.7} & 97.8 & \multicolumn{1}{c|}{98.2} & {\ul 45.3} & 73.1 & {\ul 82.7} & \multicolumn{1}{c|}{{\ul 85.5}} & \textbf{47.1} & \textbf{90.6} & \textbf{93.2} & \textbf{95.1} \\ \hline
\end{tabular}
\end{table*}

\begin{table*}[h]
\caption{Parameter analysis about different backbone. ResNet50* denotes ResNet50 with improved pre-trained weights by\cite{2021Resnet}.}
\label{tb:extra}
\centering
\begin{tabular}{cl|cccc|cccc|cccc}
\hline
\multirow{2}{*}{Clustering} & \multicolumn{1}{c|}{\multirow{2}{*}{Backbone}} & \multicolumn{4}{c|}{Market-1501} & \multicolumn{4}{c|}{MSMT17} & \multicolumn{4}{c}{VeRi-776} \\ \cline{3-14} 
 &  & mAP & R1 & R5 & R10 & mAP & R1 & R5 & R10 & mAP & R1 & R5 & R10 \\ \hline
\multirow{3}{*}{DBSCAN} & ResNet50 & 84.5 & 93.4 & 97.2 & 98.1 & 47.4 & 73.7 & 83.5 & 86.5 & 44.5 & 86.2 & 89.8 & 91.7 \\
 & ResNet50* & 84.6 & 93.5 & 97.4 & 98.0 & 52.0 & 77.0 & 86.2 & 88.8 & 44.3 & 86.5 & 90.6 & 92.7 \\
 & IBN-ResNet50 & 86.1 & 94.3 & 97.3 & 98.1 & 53.9 & 78.4 & 86.8 & 89.4 & 45.4 & 87.7 & 92.0 & 93.1 \\ \hline
\multirow{3}{*}{$K$\!-means} & ResNet50 & 85.5 & 94.0 & 97.3 & 98.1 & 34.5 & 65.2 & 75.4 & 79.4 & 46.6 & 89.8 & 92.8 & 94.3 \\
 & ResNet50* & 86.0 & 94.4 & 97.5 & 98.1 & 42.2 & 70.8 & 80.7 & 84.0 & 46.7 & 89.8 & 93.2 & 94.5 \\
 & IBN-ResNet50 & 86.7 & 94.3 & 97.5 & 98.3 & 44.3 & 72.4 & 81.8 & 84.9 & 47.0 & 90.8 & 93.3 & 94.9 \\ \hline
\multirow{3}{*}{HDC} & ResNet50 & 86.7 & 94.7 & 97.8 & 98.6 & 45.3 & 73.1 & 82.7 & 85.5 & 47.1 & 90.6 & 93.2 & 95.1 \\
 & ResNet50* & 87.2 & 95.1 & 97.7 & 98.4 & 48.2 & 75.0 & 84.3 & 87.2 & 48.2 & 90.9 & 93.8 & 95.0 \\
 & IBN-ResNet50 & 87.5 & 94.5 & 97.7 & 98.5 & 50.8 & 77.0 & 85.6 & 88.4 & 48.2 & 91.3 & 94.1 & 95.5 \\ \hline
\end{tabular}
\end{table*}

In Table \ref{tb:sota}, the proposed 3C framework is compared against state-of-the-art USL person Re-ID methods. 
 Performances surpassing all competing methods are highlighted in  \textbf{bold}, and the second-best performances are marked with {\ul underline}.
Overall, the 3C framework has the best mAP and Rank-1 results on Market-1501, and best results in all cases on MSMT17 and VeRi-776.

Compared to fully unsupervised methods, the 3C framework achieves the SOTA  mAP and Rank-1 results as identical as DCMIP \cite{ICCV23DCMIP} on Market-1501. While on MSMT17, the results of the 3C framework are rather better than those of its unsupervised competitors. Moreover, the 3C framework achieves SOTA results on VeRi-776.

In particular, compared to Cluster-Contrast \cite{ACCV22ClusterContrast} which is the baseline of many works, the 3C framework with HDC achieves 3.7\%/1.8\%, 12.3\%/11.1\% and 6.3\%/4.4\% improvement in mAP/Rank-1 accuracy on Market-1501, MSMT17 and VeRi-776, respectively.

As for camera aware methods, ICE \cite{ICCV21ICE}, PPLR \cite{CVPR22PPLR} and HCACE \cite{TMM24HCACE} follow the camera setting of CAP \cite{AAAI21CAP}. And PPLR is the best CAP-based method so far. Compared to PPLR, the 3C framework exceeds by 2.3\%/0.4\%, 3.1\%/-0.2\% and 3.6\%/2.3\% in mAP/Rank-1 accuracy on Market-1501, MSMT17 and VeRi-776, respectively.
Compared to CCL \cite{TCSVT23CCL}, which is a SOTA method among the camera aware memory update approaches, the 3C framework improves 1.4\%/0.6\%, 3.5\%/1.7\% and 4.5\%/3.6\% in mAP/Rank-1 accuracy on Market-1501, MSMT17 and VeRi-776, respectively.

Furthermore, by replacing HDC with DBSCAN, the implementation of the 3C framework reaches the SOTA results on MSMT17. This exhibits the potential of the 3C framework in unsupervised Re-ID tasks.

\subsection{Impact of backbone}
In Section \ref{sec:warm-up}, the effectiveness of the warm-up process with DBSCAN demonstrates the performance gain realised from using high quality features. To further explore the growth point of the performance of the 3C framework, extra experiments using other two improved pre-trained model weights of ResNet50* \cite{2021Resnet} and  IBN-ResNet50 \cite{2018IBN} are listed in Table \ref{tb:extra}.

The results show that with the use of more advanced backbone networks, the results of the 3C framework and its DBSCAN and $K$\!-means variants are all enhanced. In particular, the improvement of $K$\!-means, which is sensitive to the quality of features, is the most significant on MSMT17. 
Given these results, in practice, the 3C framework can be expected to have better performance by using more superior backbone networks.

\section{Conclusion}
\label{sec:conclusion}
In this paper, a confidence-guided clustering and contrastive learning (3C) framework is proposed for unsupervised person Re-ID. This 3C framework investigates three confidence degrees: the confidence of the discrepancy between samples and clusters, the confidence of the camera diversity in a cluster and the confidence of the hard sample in each cluster. In line with these three confidence degrees, a harmonic discrepancy clustering algorithm (HDC), a camera information entropy (CIE) and a confidence integrated harmonic discrepancy (CHD) are further proposed for the 3C framework. 
Experimental results demonstrate the superiority of the 3C framework and its variants, over state-of-the-art methods. In particular, it makes significant performance gains in complex scenarios, such as MSMT17 and VeRi-776.

Topics for further research include a more comprehensive study of how the 3C framework could be used for other unsupervised Re-ID tasks, given its outperformance for complex scenarios shown in this paper.  
In addition, an investigation into potential time efficiency improvement of HDC will be considered for future work.
Since CIE can be used as an indicator of the quality of model training, the underlying relationship between CIE and accuracy results is a worthwhile avenue of exploration.
Furthermore, the confidence of hard samples in CHD can be used to develop alternative sampling approach to memory updating. 
Last but not least, based on the proposed 3C framework,
potential alternative implementations with superior backbone networks in practice remain active research.

{\small
\bibliographystyle{IEEEtran}
\bibliography{references}
}

\end{document}